\pdfoutput=1

\documentclass[11pt]{article}

\usepackage[final]{acl}

\usepackage{times}
\usepackage{latexsym}

\usepackage[T1]{fontenc}

\usepackage[utf8]{inputenc}

\usepackage{microtype}

\usepackage{inconsolata}

\usepackage{graphicx}
\usepackage{algorithm}
\usepackage{algorithmic}
\usepackage{multirow}
\usepackage{pifont}
\usepackage{xcolor}
\usepackage{setspace}
\usepackage{booktabs}
\usepackage{hyperref}
\usepackage{boxedminipage}
\usepackage{amsmath}
\usepackage{amsfonts}
\usepackage{soul}
\usepackage{enumitem}
\usepackage[figuresright]{rotating}
\usepackage[most]{tcolorbox}
\usepackage{stfloats}

\definecolor{my_green}{RGB}{50,220,0}
\definecolor{my_yellow}{RGB}{255,165,0}
\definecolor{my_red}{RGB}{255, 0, 0}
\definecolor{error_red}{RGB}{241,156,153}
\definecolor{pass_green}{RGB}{185,224,165}
\definecolor{shadecolor}{RGB}{237,237,237}

\newcommand{\cmark}{\textcolor{my_green}{\ding{51}}} 
\newcommand{\xmark}{\textcolor{my_red}{\ding{55}}} 

\newcommand{\model}{\textsc{Samoyed}}
\newcommand{\dataset}{\textsc{AgentBank}}

\title{\dataset{}: Towards Generalized LLM Agents via Fine-Tuning on \\50000+ Interaction Trajectories}

\author{%
  Yifan Song$^1$,
  Weimin Xiong$^1$,
  Xiutian Zhao$^2$,
  Dawei Zhu$^1$,
  Wenhao Wu$^1$, \\
  \textbf{Ke Wang}$^3$\textbf{,}
  \textbf{Cheng Li}$^3$\textbf{,}
  \textbf{Wei Peng}$^3$\textbf{,}
  \textbf{Sujian Li}$^1$\thanks{Corresponding Authors.}\\
  $^1$National Key Laboratory for Multimedia Information Processing,\\ School of Computer Science, Peking University\quad\\
  $^2$University of Edinburgh\quad $^3$Huawei Technologies\quad\\
  \texttt{\{yfsong, lisujian\}@pku.edu.cn} \\
  \texttt{\raisebox{-0.17\height}{\includegraphics[width=0.04\textwidth]{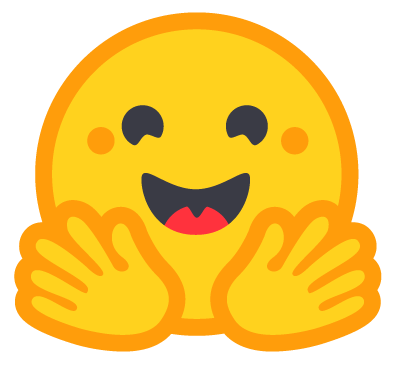}} \url{https://huggingface.co/datasets/Solaris99/AgentBank}}\\
}

\begin{document}
\maketitle

\begin{abstract}
Fine-tuning on agent-environment interaction trajectory data holds significant promise for surfacing generalized agent capabilities in open-source large language models (LLMs).
In this work, we introduce \dataset{}, by far the largest trajectory tuning data collection featuring more than 50k diverse high-quality interaction trajectories which comprises 16 tasks covering five distinct agent skill dimensions.
Leveraging a novel annotation pipeline, we are able to scale the annotated trajectories and generate a trajectory dataset with minimized difficulty bias.
Furthermore, we fine-tune LLMs on \dataset{} to get a series of agent models, \model{}.
Our comparative experiments demonstrate the effectiveness of scaling the interaction trajectory data to acquire generalized agent capabilities.
Additional studies also reveal some key observations regarding trajectory tuning and agent skill generalization.
\end{abstract}

\section{Introduction}

An agent is an entity that possesses the capability for volition, decision-making, action-taking , and, most critically, environment perception~\citep{jennings1998roadmap}.
In the realm of cognitive science, previous literature has suggested that interaction with environment derives an agent's generalized intelligence, and intelligent behavior emerges from a synergistic blend of simpler behaviors, including reasoning, programming, and game playing~\citep{brooks1991intelligence}.
The proprietary large language models (LLMs), such as GPT-3.5~\citep{chatgpt} and GPT-4~\citep{gpt4}, have demonstrated strong capabilities in instruction following, reasoning, and planning, which encourage many attempts to build autonomous agent systems utilizing LLMs as core controllers~\citep{autogpt,song2023restgpt}.
However, comprehensive evaluations have shown that the majority of open-sourced LLMs fall short in agent capabilities when compared with GPTs~\citep{liu2023agentbench,wang2023mint}.

Previous research  pointed out that learning from gold interaction trajectories, a process we term \textbf{Trajectory Tuning}, could enhance the capabilities of weaker agents~\citep{brooks1991intelligence,hussein2017imitation}.
Early studies heavily focus on specialized agents designed for particular tasks. 
Existing attempts are exemplified by \citet{chen2023fireact}, \citet{yin2023lumos}, and \citet{song2024trial}, who build agent trajectory data from teacher agents (\textit{e.g.}, GPT-4) and fine-tune open-source LLMs to improve specific agent abilities like reasoning.
Taking a step further, \citet{zeng2023agenttuning} adopt a multi-task tuning approach called AgentTuning.
However, trained on a small trajectory dataset comprising six tasks with 1.8k trajectories, \citet{zeng2023agenttuning} struggle to enhance the generalized agent capability, especially in the case of 7B and 13B models.

\begin{table*}[t!]
\centering
\resizebox{0.94\linewidth}{!}{
\begin{tabular}{lccccc}
\toprule
 & \textbf{\dataset{}} & \textbf{FireAct} & \textbf{AgentInstruct} & \textbf{Agent-FLAN} & \textbf{AgentOhana} \\
 & (this work) & \citep{chen2023fireact} & \citep{zeng2023agenttuning} & \citep{chen2024agent} & \citep{zhang2024agentohana} \\
\midrule
Number of tasks & \textbf{16} & 3 & 6 & 7 & \underline{10} \\
Number of trajectories & \textbf{51287} & 1344 & 1866 & 24703 & \underline{42600} \\
Average interaction turns & \underline{3.9} & - & \textbf{5.2} & 3.7 & 3.1 \\
No difficulty bias? & \cmark & \xmark & \xmark & \xmark & \xmark \\
Open-sourced? & \cmark & \cmark & \cmark & \cmark & \xmark \\
\midrule
Reasoning & \cmark & \cmark & \xmark & \xmark & \cmark \\
Math & \cmark & \xmark & \xmark & \xmark & \xmark \\
Programming & \cmark & \xmark & \cmark & \cmark & \cmark \\
Web & \cmark & \xmark & \cmark & \cmark & \cmark \\
Embodied AI & \cmark & \xmark & \cmark & \cmark & \cmark \\
\bottomrule
\end{tabular}
}
\caption{A comparison of \dataset{} with other datasets for agent trajectory tuning.}
 \label{tab:comparison}
\end{table*}

To explore the impacts of incorporating interaction trajectory data on agent ability generalization, we construct \dataset{}, the largest agent interaction trajectory dataset to date. \dataset{} features 16 distinct tasks across five agent skill dimensions and contains over 50,000 trajectories, each annotated with high-quality chain-of-thought (CoT) rationale for every step of action.
Leveraging a novel annotation pipeline that fully exploits the capability of LLMs, the trajectory collection process is highly scalable and adaptable to diverse agent environments.
In contrast to prior studies that have relied on successful trajectories of GPTs for training data~\citep{chen2023fireact,zeng2023agenttuning}, \dataset{} stands out with its exceptional quality and mitigated susceptibility to the \textit{difficulty bias} issue.

We further develop \model{}, a suite of models with enhanced agent capabilities, through the trajectory tuning of Llama-2~\citep{touvron2023llama} using \dataset{}.
Our evaluations on both held-in and unseen held-out tasks suggest that by fine-tuning on extensive multi-task trajectories, our models exhibit remarkable agent intelligence in comparison with untuned ones.
Specifically, \model{} outperforms GPT-3.5-Turbo on average on held-in tasks, which can be attributed to the in-domain trajectory tuning.
Furthermore, our models also demonstrate superior performance on held-out tasks, underscoring the efficacy of large-scale trajectory tuning in acquiring generalized agent capabilities.

To trace the emergence of agent capabilities generalization, we follow the initial evaluation with a systematic analysis across various dimensions.
Initially, we delineate the scaling trends of tasks alongside the quantity of trajectories.
Next, we conduct an ablation study that merges generalist instruction data and code data to examine the benefits of hybrid training. This study uncovers further enhancements in the agent capabilities and mitigates catastrophic forgetting.
Furthermore, our findings underscore the pivotal role of CoT rationale in the acquisition of generalized agent capability.

Our contributions are summarized as follows:
\begin{itemize}[leftmargin=*]
    \item The release of \dataset{}, a dataset of over 50,000 high-quality agent interaction trajectories, spanning 16 tasks across five skill dimensions. We also present a novel annotation pipeline, offering scalability and a marked reduction in difficulty bias, surpassing previous methods.
    \item The development of \model{}, the most powerful open-source LLM suite at the 7B/13B scale optimized for agent tasks. Trained through trajectory tuning, \model{} demonstrates exceptional performance, showcasing transferable agent intelligence on unseen tasks.
    \item We conduct comprehensive experiments and in-depth analysis on agent intelligence acquisition, including the relations with instruction following and code capability, scaling law of interaction trajectories, and the effectiveness of training with CoT.
\end{itemize}

\begin{figure*}
    \centering
    \resizebox{\linewidth}{!}{
    \includegraphics{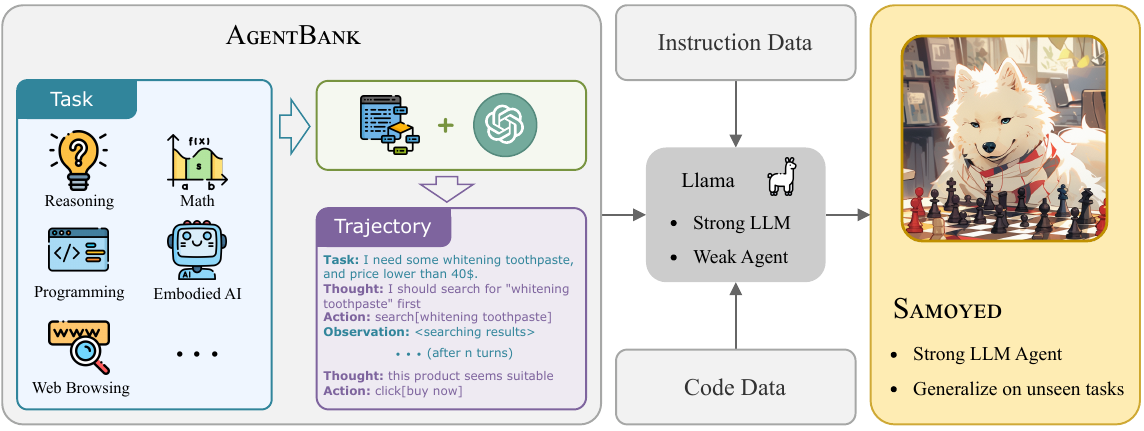}
    }
    \caption{Overview of the construction process of \dataset{} and the training procedure of \model{}}
    \label{fig:main}
\end{figure*}

\section{Related Work}

\subsection{Instruction Tuning}

Instruction tuning is a simple yet powerful approach to align LLMs with human preferences~\citep{zhang2023instruction}.
Previous studies have primarily focus on improving general-purpose instruction following capabilities of LLMs.
FLAN series~\citep{wei2021finetuned,chung2022scaling}, T0~\citep{sanh2021multitask}, and NaturalInstruction~\citep{wang2022super} scale up the instruction datasets to activate the generalized instruction following capabilities of LLMs.
More recently, utilizing synthetic instruction following data distilled from GPTs to align open-source LLMs has also been proposed~\citep{alpaca,vicuna2023}.
Furthermore, multiple works have shown the promise of instruction tuning in enhancing the specialized abilities of LLMs, such as math~\citep{yu2023metamath,yue2023mammoth}, reasoning~\citep{lee2023platypus}, and agent tasks~\citep{chen2023fireact,zeng2023agenttuning}.

\subsection{LLM-based Agent}
Modern LLMs have demonstrated various emergent abilities that encourage researchers to build agent systems based on LLMs.
ReAct~\citep{yao2022react} combines CoT reasoning with agent actions to accomplish tasks such as QA.
AutoGPT~\citep{autogpt} harnesses LLMs as the core controllers to constitute powerful agent frameworks capable of solving real-world complex problems.
While advanced proprietary models exampled by GPT-3.5/4 have shown strong performances on agent tasks, their open-source counterparts still lag far behind \citep{liu2023agentbench,wang2023mint}.
In response, recent studies including FireAct \citep{chen2023fireact}, AgentTuning \citep{zeng2023agenttuning} and AgentOhana~\citep{zhang2024agentohana} collect agent trajectory data from teacher agents (\textit{e.g.}, GPT-4) and fine-tune open-source LLMs (\textit{e.g.}, Llama series) with the data.
However, limited by the number of tasks and expert trajectories, existing research has not yet exhaustively explored whether open-source LLMs can acquire generalized agent abilities, a gap that this study aims to bridge.

\section{Preliminary}

\subsection{Agent Task Formulation}
Given an agent task described by the instruction $u$, an LLM agent generates an action $a_1$ based on its policy.
Next, an environment receives the action, transfers to a new latent state, and provides an observation $o_i$ in natural language format.
Subsequently, the agent generates another action for the next step, $a_{i+1}$, and repeats this circle of interaction with the task environment until either the task is completed or the maximum number of steps is reached.
This ``conversation'' between the agent with the environment is denoted as the interaction trajectory $\left(u,a_1,o_1,...,a_n\right)$.
Finally, a final reward $r\in\left[0,1\right]$ is returned depending on the task completion status.

Chain-of-Thought (CoT) ~\citep{wei2022chain,kojima2022large} is an effective approach to enhance the inferential capabilities of LLMs by a step-by-step reasoning process.
We employ ReAct~\citep{yao2022react} as the agent tasking framework, which outputs rationale before the action. 

\subsection{Challenges in Trajectory Collection}

Previous works~\citep{chen2023fireact,zeng2023agenttuning} have employed GPT-4 as teacher agents to interact with the environment and collect successful interaction trajectories.
To ensure the quality of generated data, a failure filtering mechanism is used to remove the cases where GPT failed.
However, this \textbf{GPT-exploration} pipeline automates the trajectory construction at some significant cost.

\paragraph{Hard to Scale-Up}
The quality of data is essential for agent training, and training with failure trajectories will lead to performance degradation~\citep{zeng2023agenttuning}.
Therefore, scaling up this process to a larger trajectory amount is challenging due to the low success rate of GPT-4.
For instance, AgentInstruct~\citep{zeng2023agenttuning} discards more than $90\%$ generated trajectories due to GPT failures.

\paragraph{Difficulty Bias}
Even worse, GPT-exploration pipelines will inevitably introduce difficulty bias to the final training data.
Essentially, a trajectory filtering strategy can be regarded as grouping the instances based on whether GPT is capable of solving them.
Discarding failed trajectories leads to a skewed distribution of ``difficulty'', resulting in a training set with much easier instances than those in the test set.
This violation of the \textit{i.i.d.} assumption may hurt the generalization ability of the trained agents.
In Appendix~\ref{app:bias}, we conduct an experiment to show this bias.

\section{\dataset{}}
\label{sec:method}

In response to the challenges of previous trajectory collection pipeline, we propose a new trajectory annotation pipeline and construct \dataset{} trajectory dataset.

\begin{table*}[t!]
\centering
\resizebox{\linewidth}{!}{
\begin{tabular}{llccccc}
\toprule
\textbf{Skill Dim.} & \textbf{Task} & \textbf{Action Space} & \textbf{Tool} & \textbf{\#Inst.} & \textbf{Avg. Turns} & \textbf{Action Annotation} \\
\midrule
\multirow{3}{*}{Reasoning}
& HotpotQA~\citep{yang2018hotpotqa} & Continuous & Search & 4273 & 3.1 & Explore \\
& StrategyQA~\citep{geva2021did} & Continuous & Search & 1267 & 3.6 & Explore \\
& TriviaQA~\citep{joshi2017triviaqa} & Continuous & Search & 4134 & 2.5 & Explore \\
\midrule

\multirow{3}{*}{Math}
& GSM8K~\citep{cobbe2021training} & Continuous & Calculator & 7471 &  4.5 & Reformat \\
& MathQA~\citep{amini2019mathqa} & Continuous & Python & 4000 & 2.0 & Explore \\
& MATH~\citep{hendrycks2021measuring} & Continuous & Python, Wiki & 2312 & 2.5 & Explore \\
\midrule

\multirow{4}{*}{Programming}
& IC-SQL~\citep{yang2023intercode} & Continuous & MySQL & 4540 & 4.8 & Explore+Answer Force \\
& APPS~\citep{hendrycks2021measuring} & Continuous & Python & 4408 & 1.0 & Reformat \\
& HumanEval~\citep{chen2021evaluating} & Continuous & Python & 134 & 2.7 & Explore+Answer Force \\
& MBPP~\citep{austin2021program} & Continuous & Python & 608 & 2.2 & Explore+Answer Force \\
\midrule

\multirow{3}{*}{Web}
& Mind2Web~\citep{deng2023mind2web} & Discrete & - & 7770 & 1.0 & Reformat \\
& WebArena~\citep{zhou2023webarena} & Discrete & - & 657 & 1.0 & Reformat \\
& WebShop~\citep{yao2022webshop} & Discrete & - & 5315 & 3.4 & Explore \& Reformat \\
\midrule

\multirow{3}{*}{Embodied}
& ALFWorld~\citep{shridhar2020alfworld} & Discrete & - & 3554 & 10.1 & Reformat \\
& RoomR~\citep{RoomR} & Discrete & - & 300 & 30.2 & Search+Reformat \\
& IQA~\citep{gordon2018iqa} & Discrete & - & 1627 & 28.4 & Search+Reformat \\
\midrule

& Total (\dataset{}) & - & - & 51287 & 3.9 & - \\
\bottomrule
\end{tabular}
}
\caption{Overview of \dataset{} dataset. It compiles 16 agent tasks covering 5 skill dimensions, formulating the largest interaction trajectory dataset. ``Inst.'' and ``Traj.'' refer to instruction and interaction trajectory.}
 \label{tab:superagent}
\end{table*}

\subsection{Task and Instruction Collection}

A generalized agent needs to possess a wide range of capabilities across various dimensions. To this end,
as shown in Table \ref{tab:superagent}, we curate 16 publicly available agent datasets to lay the foundation of \dataset{} and categorize specific tasks into five skill dimensions: reasoning, math, programming, web navigation, and embodied tasks.
Additionally, some tasks aggregated in \dataset{} involve the usage of external tools, such as search engine, calculator, and code interpreter, as the ability to effectively operate tools is also a crucial aspect for generalized agents.
From the perspective of action space, tasks in \dataset{} can be classified into two types: those with a \textit{continuous} action space (including natural language and code) and those with a predefined \textit{discrete} action space.
Our dataset also covers a broad range of interaction turns, ranging from 1 to 30.
Note that some tasks are originally evaluated in a single-turn QA style, such as HotpotQA~\citep{yang2018hotpotqa} and MATH~\citep{hendrycks2021measuring}.
Following \citet{wang2023mint}, we modify these datasets to accommodate multi-turn interaction environments with tool usage.

Since most of the original benchmarks have a training set, we use them to construct our dataset.
To balance data sources, we down-sample some tasks which have a huge training set.
See Appendix \ref{app:dataset} for detailed descriptions of each dataset.

\subsection{Action Annotation}
\label{sec:annotation}

To tackle the challenges in trajectory collection, unlike previous methods that generate action and CoT simultaneously, we separate the annotation of gold actions and their corresponding rationales, fully leveraging of the capability of LLMs.

Specifically tailored to the specific nature of different tasks, our approach involves several techniques to obtain high-quality action sequences accordingly.

\paragraph{Answer Forcing} 
For tasks characterized by a continuous natural language or code action space, such as IC-SQL, we introduce an answer forcing action annotation strategy as an extension to GPT-exploration pipeline.
This strategy aims to mitigate the bias introduced by failure filtering.
Initially, we use GPT-4 to interact with the environment and gather interaction trajectories.
For failed trajectories, rather than directly discarding them, we prompt GPT with the failed trajectory and the gold final answer to generate a new interaction trajectory.
Then we validate the correctness of new trajectories by executing the actions within real agent environments.
This answer forcing process is used in an iterative manner to re-annotate failure trajectories and generate a substantial number of gold action sequences.
See Appendix~\ref{app:prompt_ann} for the re-annotation prompt.

\paragraph{Heuristic Action Search}
For tasks with a discrete action space, exemplified by embodied AI tasks~\citep{RoomR,gordon2018iqa}, we are able to access both the environment's source code and its complete execution state.
Leveraging this access, we employ the heuristic depth-first search algorithm to efficiently get the optimal action sequences.

\paragraph{Reformat}
Some tasks have already provided official solving trajectories.
For instance, GSM8K~\citep{cobbe2021training} offers ground-truth intermediate reasoning steps.
For these tasks, following~\cite{yin2023lumos}, we exploit GPT as a style transfer tool to transform reasoning process into agent interaction action sequences.

\subsection{Rationale Annotation}

Give the instructions and gold action sequences, we directly prompt GPT to generate the corresponding CoT rationale of each action step.
Since providing explanation for gold actions is relatively easy task, we employ GPT-3.5-Turbo as the primary LLM in the rationale annotation process.
The rationale generation prompt is shown in Appendix~\ref{app:prompt_ann}.
We also compare rationales generated by different LLMs in Appendix~\ref{app:rationale_compare}.

For tasks with a huge number of instructions and GPT-4 have a high success rate, such as StrategyQA~\citep{geva2021did} and WebShop~\citep{yao2022webshop}, we directly use the GPT-exploration pipeline as \citet{zeng2023agenttuning}.

The overview of \dataset{} is shown in Table \ref{tab:superagent}.
See the Appendix \ref{app:dataset} for more details about the annotation process of each task.
A human evaluation assessing the quality of our dataset can be found in Appendix \ref{app:quality}.

\section{Train \model{} with \dataset{}}

\begin{table}[t]
\centering
\resizebox{\linewidth}{!}{
\begin{tabular}{lcccc}
\toprule
\textbf{Task} & \textbf{Skill Dim.} & \textbf{\#Inst.} & \textbf{Metric} \\
\midrule
\multicolumn{4}{c}{\textit{Held-in Tasks}} \\
\midrule
HotpotQA~\citep{yang2018hotpotqa} & Reasoning & 100 & Exact Match \\
StrategyQA~\citep{geva2021did} & Reasoning & 100 & Exact Match \\
GSM8K~\citep{cobbe2021training} & Math & 100 & Exact Match \\
MATH~\citep{hendrycks2021measuring} & Math & 100 & Exact Match \\
IC-SQL~\citep{yang2023intercode} & Programming & 100 & Avg. Reward \\
MBPP~\citep{austin2021program} & Programming & 100 & Success Rate \\
Mind2Web~\citep{deng2023mind2web} & Web & 1173 & Step SR \\
WebShop~\citep{yao2022webshop} & Web & 200 & Avg. Reward \\
ALFWorld~\citep{shridhar2020alfworld} & Embodied & 134 & Success Rate \\
\midrule
\multicolumn{4}{c}{\textit{Held-out Tasks}} \\
\midrule
Bamboogle~\citep{press2022measuring} & Reasoning & 126 & Exact Match \\
TheoremQA~\citep{chen2023theoremqa} & Math & 100 & Exact Match \\
IC-Bash~\citep{yang2023intercode} & Programming & 200 & Avg. Reward \\
MiniWoB++~\citep{kim2023language} & Web & 460 & Success Rate \\
ScienceWorld~\citep{wang2022scienceworld} & Embodied & 270 & Avg. Reward \\

\bottomrule
\end{tabular}
}
\caption{The held-in and held-out tasks used to evaluate the agent capabilities of different LLMs.}
\label{tab:test_tasks}
\end{table}

To initialize the training of \model{}, we formulate agent interaction trajectories in \dataset{} into a chatbot-style schema $\left(u,a_1,o_1...,a_i,o_i,...,a_n\right)$, where $u$ is the task instruction, $o_i$ and $a_i$ denote the observation from the task environment and the corresponding action with rationale generated by the agent in the $i$-th round.
During the training process, we feed the entire interaction trajectory into a decoder-only LLM, where only the auto-regressive loss on tokens of ground-truth responses $Y=\{a_1,...,a_n\}$ is counted.
We mask all tokens belonging to the instruction and observations from the environment to prevent them from loss computation.
Concretely, the loss function is defined as:
\begin{equation}
    \mathcal{L}=-\sum_j\log p_\theta(t_j|t_{<j})\times \mathbf{1}(t_j\in Y),
\end{equation}
where $t_j$ denotes the $j$-th input token and $\mathbf{1}$ is the indicator function.

Recent studies~\citep{yang2024if,zeng2023agenttuning} suggest that hybrid training with generalist instruction data and code data may improve the generalized ability of LLM agents.
Following them, we adopt a mixture of \dataset{} $\mathcal{D}_{\mathrm{agent}}$, the general domain instruction dataset $\mathcal{D}_{\mathrm{general}}$, and the code dataset $\mathcal{D}_{\mathrm{code}}$ for fine-tuning.
We perform detailed ablation experiments to explore the effectiveness of generalist and code data in Section~\ref{sec:sharegpt}.

\begin{table*}[t]
\centering
\resizebox{\textwidth}{!}{
\begin{tabular}{l|ccccc|c|ccccc|c}
\toprule

\multirow{2}{*}{\textbf{Model}} & \multicolumn{6}{c|}{\textbf{Held-in Tasks}} & \multicolumn{6}{c}{\textbf{Held-out Tasks}} \\
\cmidrule(l){2-7} \cmidrule(l){8-13} 
& Reason & Math & Program & Web & Embodied & Avg. & Reason & Math & Program & Web & Embodied & Avg. \\
\midrule

\multicolumn{13}{c}{\textit{Closed-Source Model}} \\
\midrule
GPT-4  & 61.6 & 73.0 & 54.9 & 40.6 & 77.8 & 59.8 & 41.6 & 51.0 & 69.4 & 69.4 & 36.4 & 53.6 \\ 
GPT-3.5-Turbo & 41.0 & 41.5 & 51.2 & 42.0 & 10.5 & 40.2 & 32.0 & 32.0 & 54.8 & 66.7 & 21.2 & 41.3 \\ 
\midrule

\multicolumn{13}{c}{\textit{7B Open-Source Model}} \\
\midrule
Llama-2-7B-Chat  & 4.0 & 7.5 & 2.5 & 13.9 & 0.0 & 6.2 & 4.0 & 8.0 & 7.0 & 0.4 & 7.8 & 5.5  \\ 
Vicuna-7B  & 29.0 & 2.0 & \underline{19.0} & 24.2 & 6.0 & 17.1 & 8.8 & \underline{14.0} & 19.0 & 18.2 & 12.8 & 14.6  \\
CodeLlama-7B  & 3.5 & 3.5 & 1.5 & 24.8 & 0.0 & 7.4 & 1.0 & 13.0 & 21.8 & \textbf{41.3} & 5.5 & 16.5  \\ 
AgentLM-7B  & 29.5 & 10.0 & 12.0 & \textbf{37.2} & \underline{63.4} & 26.7 & 19.2 & 13.0 & 50.5 & 13.5 & 13.3 & 21.9 \\ 
Agent-FLAN-7B & \underline{31.0} & \underline{10.5} & 13.1 & 35.4 & \textbf{65.3} & \underline{27.3} & \underline{22.2} & 11.0 & \underline{53.1} & 17.9 & \underline{14.1} & \underline{23.7} \\
\midrule
\model{}-7B  & \textbf{48.0} & \textbf{30.5} & \textbf{41.6} & \underline{36.4} & 61.2 & \textbf{41.6} & \textbf{32.0} & \textbf{18.0} & \textbf{59.2} & \underline{24.2} & \textbf{14.2} & \textbf{29.5} \\ 
\midrule

\multicolumn{13}{c}{\textit{13B Open-Source Model}} \\
\midrule
Llama-2-13B-Chat  & 12.5 & 10.5 & 8.2 & 11.2 & 0.0 & 9.4 & 9.6 & 11.0 & 33.0 & 17.6 & 7.3 & 15.7 \\ 
Vicuna-13B  & 25.5 & 6.5 & \underline{30.4} & 34.2 & 2.2 & 21.7 & \underline{24.8} & \underline{17.0} & 37.0 & 34.2 & \underline{14.8} & \underline{25.6} \\ 
CodeLlama-13B  & 13.5 & \underline{18.5} & 5.1 & 15.3 & 0.0 & 11.7 & 6.4 & 16.0 & 11.1 & \textbf{46.5} & 5.5 & 17.1 \\ 
AgentLM-13B  & \underline{38.0} & 13.5 & 22.8 & \underline{38.1} & \underline{52.2} & \underline{30.8} & 20.8 & 13.0 & \underline{46.6} & 21.6 & 14.6 & 23.3 \\ 
\midrule
\model{}-13B  & \textbf{54.5} & \textbf{38.5} & \textbf{55.4} & \textbf{40.9} & \textbf{72.4} & \textbf{50.1} & \textbf{35.0} & \textbf{23.0} & \textbf{62.4} & \underline{38.9} & \textbf{18.4} & \textbf{35.5} \\ 

\bottomrule
\end{tabular}
}
\caption{Performance comparison of \model{} and baseline LLMs on held-in and held-out tasks. Due to the space constraint, we group the held-in tasks according to the skill dimensions and report the average scores. The top-2 best of each model group are highlighted in \textbf{bold} and \underline{underlined} respectively. See Appendix \ref{app:main} for complete results.}
\label{tab:main-results}
\end{table*}

\section{Experiments}

\subsection{Experimental Setup}

\paragraph{Base LLMs and Baselines}
We use several LLMs to conduct experiments, including Llama-2-Chat~\citep{touvron2023llama}, CodeLlama~\citep{roziere2023code}, Mistral~\citep{jiang2023mistral}, and Llama-3-Instruct~\citep{llama3}.
However, since most baselines, including AgentLM~\citep{zeng2023agenttuning} and Agent-FLAN~\citep{chen2024agent} are tuned from Llama-2-Chat, we mainly use Llama-2-Chat as our base model for a fair comparison.
Due to our limited resources, we use 7B and 13B models for our experiments, leaving the comparison at a larger scale (\textit{e.g.}, Lemur-70B~\citep{xu2023lemur} and xLAM-8$\times$7B~\citep{zhang2024agentohana}) for the future work.
We also select GPT-3.5-Turbo~\citep{chatgpt} and GPT-4~\citep{gpt4} as strong baselines.
For all LLMs, the decoding temperature is set to 0 for the most deterministic generation.

\paragraph{Training Setup}
We use AdamW optimizer with a learning rate of 5e-5 and a cosine scheduler.
The models are trained for 3 epochs with 3\% warm-up steps.
The batch size is set to 128 and the sequence length is 2048.
We choose ShareGPT\footnote{\href{https://sharegpt.com/}{https://sharegpt.com/}} as the generalist instruction data, and Evol-CodeAlpaca~\citep{luo2023wizardcoder} as the code data.
The mixture ratio of $\mathcal{D}_\mathrm{agent}$, $\mathcal{D}_{\mathrm{general}}$, and $\mathcal{D}_{\mathrm{code}}$ is $80\%$, $10\%$, $10\%$.
A corresponding data contamination analysis can be found in Appendix \ref{app:contam}.
All experiments are conducted on 8 NVIDIA A100 80G GPUs.
We use FastChat~\citep{zheng2023judging} and PyTorch FSDP~\citep{paszke2019pytorch} for efficient training.

\paragraph{Held-in/out Tasks}

In an effort to balance the reliability and efficiency of the evaluation, we select nine tasks from \dataset{} to form the held-in test set.
For tasks with a huge test set, following \citet{wang2023mint}, we randomly sample a subset from the original test set.
To evaluate the generalized agent intelligence of \model{}, we additionally compile five unseen held-out tasks that do not exist in \dataset{} but still fall into the five skill dimensions of a foundation agent.
The held-in and held-out evaluation tasks used in the experiments are listed in Table~\ref{tab:test_tasks}.
For all evaluated tasks, 1-shot in-context example is provided in prompts.
We use average scores on held-in/out tasks to measure the overall capability of different agents.
We also report the results on AgentBench~\citep{liu2023agentbench}, another agent benchmark, in Appendix \ref{app:agentbench}.

\subsection{Main Results}

\begin{figure}[t]
    \centering
    \includegraphics[width=0.92\linewidth]{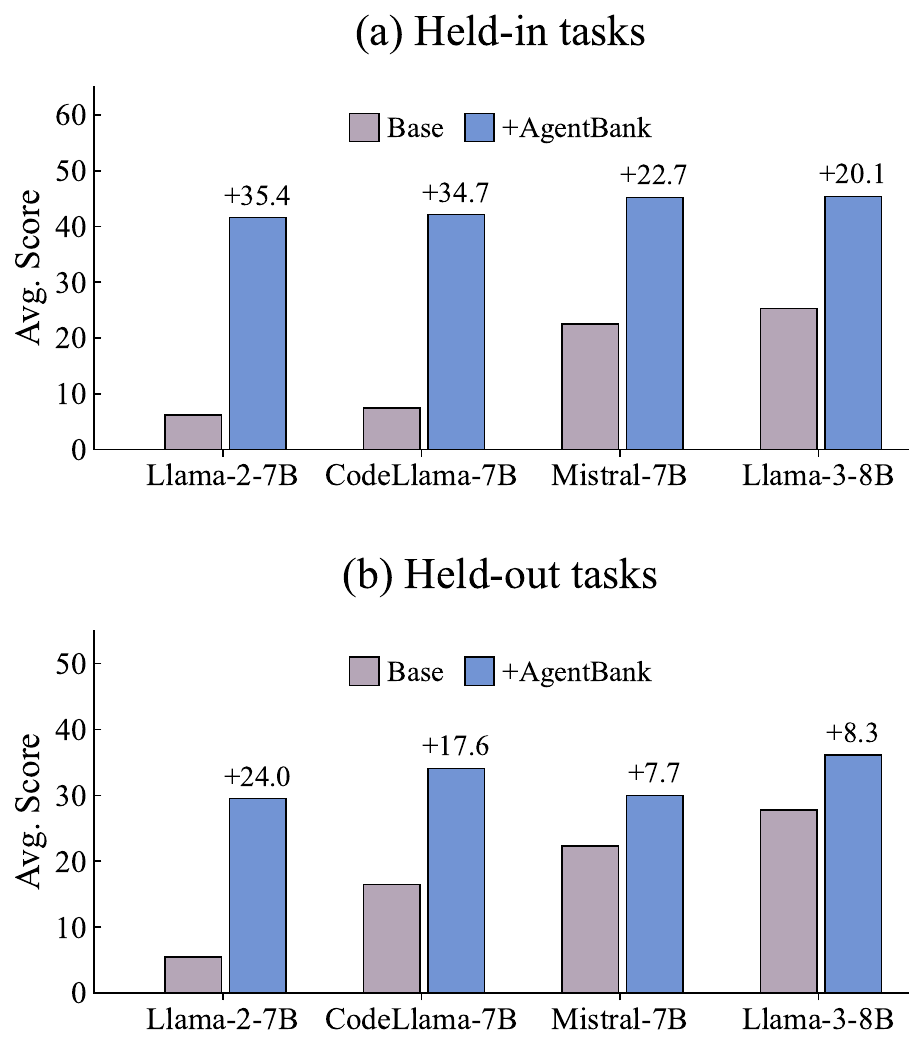}
    \caption{
    The results of different base models. ``Base'' denotes untrained LLMs. ``+SuperAgent'' denotes models after training on \dataset{}.
    }
    \label{fig:base_model}
\end{figure}

Table \ref{tab:main-results} shows the results of different models on held-in and held-out tasks.
Due to the space constraint, we grouped the held-in tasks according to skill dimensions and report the average scores.
In Figure \ref{fig:base_model}, we show the results of trajectory tuning on different base LLMs.

\paragraph{Massive trajectory tuning enables generalization to unseen tasks}
The performance of \model{} has a remarkable improvement on held-out unseen tasks, which demonstrates a substantial boost in agent capabilities through large-scale trajectory tuning.
Surprisingly, \model{}-7B exhibits an even greater enhancement compared to \model{}-13B.
Our models also outperform AgentLM and Agent-FLAN which are tuned on less trajectories, demonstrating the effectiveness of scaling up the tuning trajectories.

\paragraph{Comparison among baselines}
The experiment yields several noteworthy model-wise observations.
We find that CodeLlama, benefiting from code pretraining, excels in web browsing tasks. 
Vicuna exhibits strong abilities through fine-tuning on generalist instruction data, demonstrating impressive performance on both held-in/out tasks.
Remarkably, the performance of Vicuna-13B even surpasses AgentLM-13B.
It is important to highlight that AgentLM's training set comprises $80\%$ generalist instruction data, suggesting that the held-out task performance of AgentLM largely comes from the enhanced capability of instruction following.

\paragraph{Effectiveness of trajectory tuning on different base models}
As illustrated in Figure \ref{fig:base_model}, after large-scale trajectory tuning, all LLMs yield significant performance improvements on held-in and held-out tasks.
We also notice some interesting outcomes.
CodeLlama's superior performance indicates that code training can enhance agent capabilities.
As for Mistral and Llama-3, although fine-tuning on \dataset{} also yields improvements, the performance gain is relatively modest compared with the substantial improvement seen on Llama-2.
This finding indicates that weaker LLMs may benefit more from massive trajectory tuning than their stronger counterparts.

\section{Further Analysis}

\subsection{Scaling Trends of Generalization}

\begin{figure}[t]
    \centering
    \includegraphics[width=\linewidth]{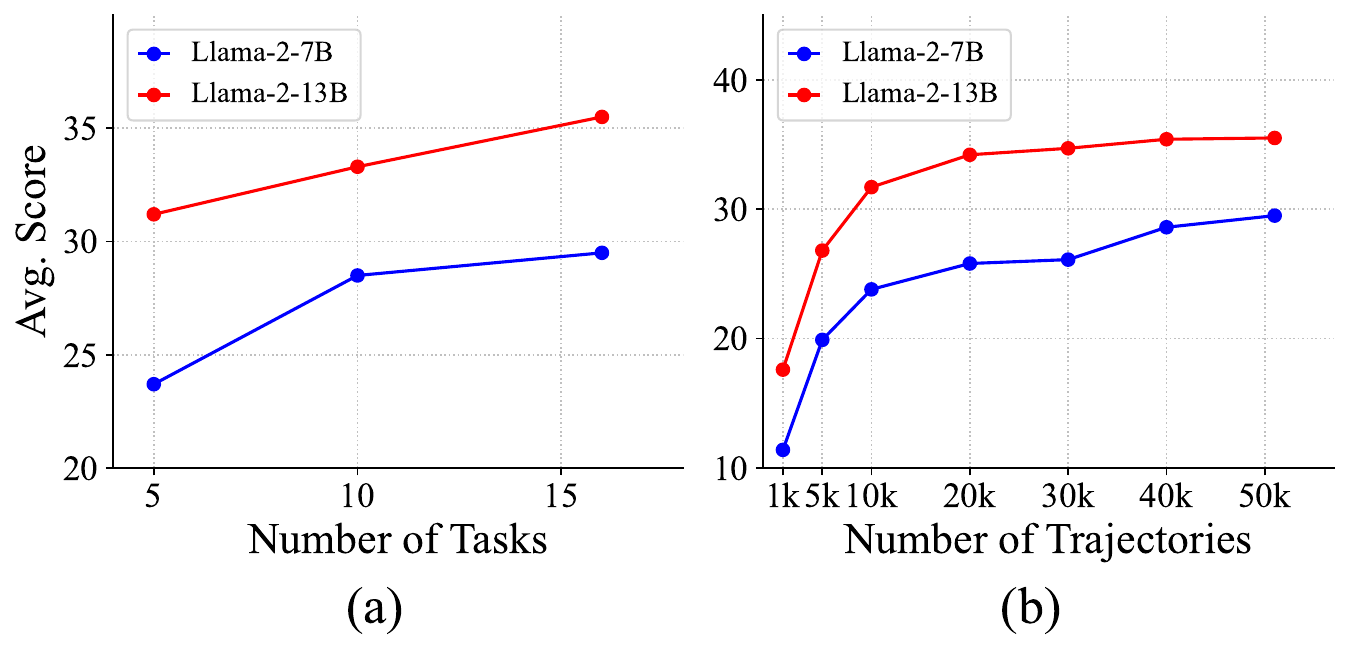}
    \caption{
    Scaling trends of the number of tasks and interaction trajectories.
    }
    \label{fig:scale}
\end{figure}

We investigate the generalization performance of trajectory tuning with respect to two scaling factors: the number of training tasks and the number of training trajectories.
Figure \ref{fig:scale} illustrates the performance changes on held-out tasks when scaling each of these factors.

To explore the impact of task scaling, we modify the number of tasks in each skill dimension while ensuring that the skill coverage of the subsets remains consistent.
We observe that increasing the number of tasks used for training results in improved performance on held-out tasks.
This finding suggests that by scaling the number of distinct tasks for trajectory tuning, the model can enhance its generalized agent capabilities.

As shown in Figure \ref{fig:scale}b, a comparison between the performance using 1k trajectories and that with 50k+ cases reveals a marked decrease in the generalized ability of the agent, highlighting the importance of scaling the amount of interaction data for better performance.
However, the trajectory of performance improvement is gradually plateauing, particularly noticeable with the 13B model, suggesting the necessity for more advanced agent training techniques beyond SFT.

\subsection{The Effect of Data Mixture}
\label{sec:sharegpt}

\begin{figure}[t]
    \centering
    \includegraphics[width=\linewidth]{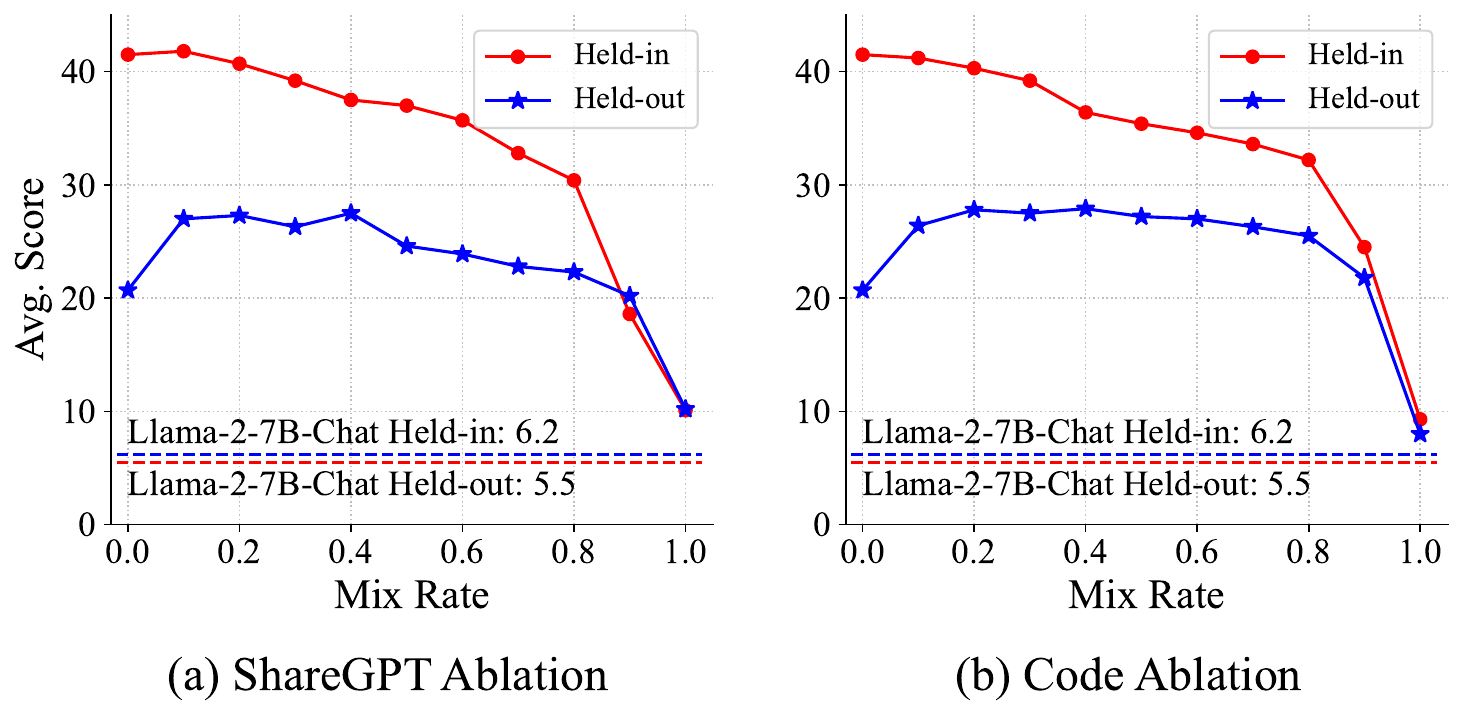}
    \caption{
    Ablation study on data mixture.
    }
    \label{fig:sharegpt}
\end{figure}

\paragraph{Mixture Training leads to better generalization.}
When training \model{}, we mix $10\%$ generalist instruction data and $10\%$ code data.
Here we conduct ablation study to investigate the effect of mixture training.
Specifically, we vary the mixture ratio of ShareGPT and code data and train Llama-2-7B-Chat for 1000 steps.
As shown in Figure~\ref{fig:sharegpt}a, a relatively low proportion of generalist data leads to improved agent performance on unseen tasks.
Nevertheless, as the amount of generalist data continues to increase, the performance on held-out tasks dramatically degrades.
Moreover, disagreed with \citet{zeng2023agenttuning} who find that training with only interaction trajectory data will lead to performance degradation on held-out tasks, \model{} trained on solely \dataset{} shows performance improvement on held-out tasks instead.

The ablation on code data also shows a lower ratio of code data will benefit the generalization ability of the agents.
Code data, comprising standard syntax and logical abstraction, has the potential to enhance the planning and decision-making capabilities of LLM agents~\citep{yang2024if}.

\begin{table}[t]
\centering
\tabcolsep=4.8pt
\resizebox{\linewidth}{!}{
\begin{tabular}{lccccc}
\toprule
\textbf{Model} & \textbf{Reason} & \textbf{Math} & \textbf{Program} & \textbf{Web} & \textbf{Embodied} \\
\midrule
Llama-2-7B-Chat & 4.0 & 8.0 & 7.0 & 0.4 & 7.8 \\
+\dataset{} & \textbf{32.0} & \textbf{18.0} & 59.2 & 24.2 & 14.2 \\
\midrule
CodeLlama-7B & 1.0 & 13.0 & 21.8 & 41.3 & 5.5 \\
+\dataset{} & 29.6 & 16.0 & \textbf{67.7} & \textbf{42.2} & \textbf{14.8} \\
\bottomrule
\end{tabular}
}
\caption{The held-out task performance of Llama-2 and CodeLlama.}
\label{tab:codellama}
\end{table}

\paragraph{Code pretraining benefits web tasks.}
As a medium between humans and computers, code translates high-level goals into executable steps, featuring standard syntax, logical consistency, and abstraction.
To further analyse the effect of code training, in Table \ref{tab:codellama}, we compare the distinctions between agents based on Llama-2-Chat and CodeLlama.
Unsurprisingly, due to its extensive code training, CodeLlama demonstrates excellent performance in programming tasks.
Training with extensive interaction trajectories can further elevate its coding proficiency.
Additionally, CodeLlama shows exceptional competence in web navigation tasks, likely attributed to the abundance of web pages present in its pretraining datasets.

\paragraph{Mixture training alleviates catastrophic forgetting.}
Supervised fine-tuning LLMs on downstream tasks will lead to catastrophic forgetting on general capabilities.
Here, we select three widely used benchmarks, MMLU~\citep{hendrycks2020measuring}, MT-Bench~\citep{zheng2023judging}, AlpacaEval 2~\citep{alpaca_eval}, to evaluate the general capabilities of the trained agents.
As shown in Table~\ref{tab:cf}, since the agent trajectory often presented in specific ReAct formats, the models are easily to get overfitting on this style when training solely on agent data.
Simply incorporating generalist instruction data during training proves to be an effective strategy in mitigating catastrophic forgetting.

\begin{table}[t]
\centering
\resizebox{\linewidth}{!}{
\begin{tabular}{lccc}
\toprule
\textbf{Model} & \textbf{MMLU} & \textbf{MT-Bench} & \textbf{AlpacaEval 2} \\
\midrule
Llama-2-7B-Chat & 48.3 & 6.2 & 5.4 \\
\midrule
\model{}-7B & 47.7 & 6.1 & 5.0 \\
w/o ShareGPT & 23.1 & 2.6 & 1.9 \\
w/o Code & 48.1 & 5.9 & 5.1 \\
\bottomrule
\end{tabular}
}
\caption{Performance on general tasks.}
\label{tab:cf}
\end{table}

\subsection{The Effect of CoT Rationale}

Chain-of-Thought (CoT) plays an vital role in LLM reasoning and planning~\citep{wei2022chain,kojima2022large}.
In our experiments, agents are trained with GPT-generated rationales for each action step and are deployed under ReAct framework~\citep{yao2022react}.
In this section, we conduct an ablation study to examine the effectiveness of CoT.

As shown in Table~\ref{tab:cot}, when it comes to held-in tasks, training without rationales has a minimal impact on performance.
Mistral-based agent without CoT even slightly surpasses the one with CoT.
Nonetheless, for unseen held-out tasks, training without rationale results in a significant performance decline.
Explanation traces provide a detailed step-by-step thought processes, enabling agents to learn from the underlying and planning process~\citep{mukherjee2023orca}.
Moreover, without rationale, the agents tend to mimic the style and get overfitting on held-in tasks.

\begin{table}[t]
\centering
\resizebox{\linewidth}{!}{
\begin{tabular}{lccc}
\toprule
\textbf{Base Model} & \textbf{w/ CoT?} & \textbf{Held-In} & \textbf{Held-Out} \\
\midrule
\multirow{2}{*}{Llama-2-7B-Chat} & \cmark & 41.6 & 29.5 \\
 & \xmark & 41.2 & 22.8 \\
\midrule
\multirow{2}{*}{Mistral-7B} & \cmark & 45.2 & 30.0 \\
 & \xmark & 45.5 & 27.5 \\
 \midrule
\multirow{2}{*}{Llama-3-8B-Instruct} & \cmark & 45.4 & 36.1 \\
 & \xmark & 43.6 & 31.8 \\
\bottomrule
\end{tabular}
}
\caption{Ablation study on CoT rationale.}
\label{tab:cot}
\end{table}

\subsection{Skill-Level Transfer}

\begin{figure}[t]
    \centering
    \includegraphics[width=0.75\linewidth]{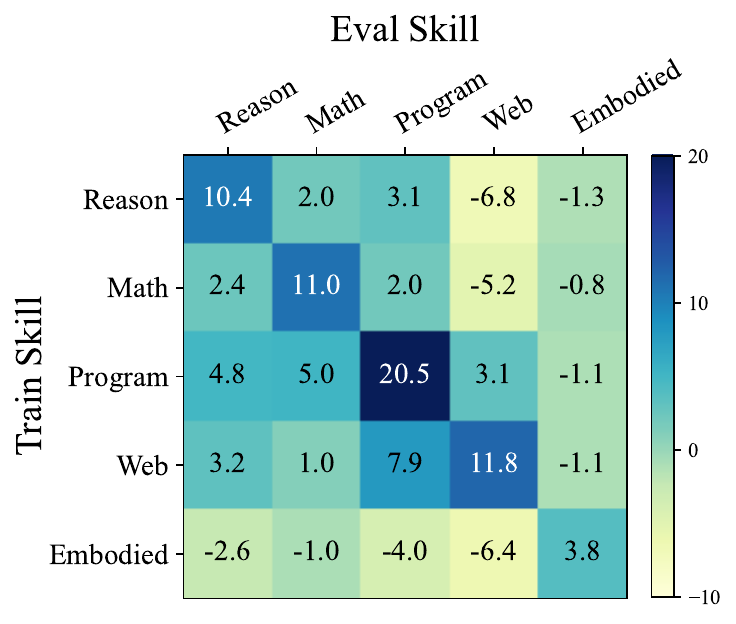}
    \caption{
    Heatmap of skill-level capability transfer. We plot the relative improvements over training on generalist instruction and code data.
    }
    \label{fig:skill}
\end{figure}

To explore the potential transferability across different agent skills, we fine-tune Llama-2-Chat on held-in tasks corresponding to a specific agent skill and evaluate on held-out tasks.
The compared baseline is fine-tuning Llama-2-Chat using a mixture of generalist instruction and code data.
All models are trained for 300 steps to ensure a fair study.

As depicted in Figure \ref{fig:skill}, most skills, with the exception of embodied skill, exhibit the ability to transfer across different skill dimensions.
This can be attributed to the unified agent interaction format in \dataset{}.
The transferability of programming and web tasks further confirms the findings from Section \ref{sec:sharegpt}.
Notably, embodied AI skill is particularly challenging, for it receives negative impact from all other skills.

\section{Conclusion}

In this work, we explore the acquisition of generalized agent capabilities through fine-tuning open-source LLMs on massive interaction trajectories.
We introduce by far the largest interaction trajectory dataset \dataset{}, comprising over 50k trajectories that encompass 16 tasks across five distinct agent skill dimensions.
Building upon \dataset{}, we fine-tune Llama-2 to develop \model{}, an open-source LLM series specialized for agent tasks.
Evaluations on both held-in and held-out tasks show that \model{} significantly outperforms strong baselines in terms of generalized agent capabilities.
Comprehensive analysis also reveals the effectiveness of data mixture and plots the scaling law of trajectories.
We hope this work to serve as a catalyst for further exploration in the development of more powerful agents.

\section*{Limitations}

We conclude the limitations of this work as follows:
\begin{itemize}[leftmargin=*]
    \item Due to the resource constraints, we only conduct experiments and analysis on 7B and 13B models. The extent to which larger models can benefit from large-scale trajectory tuning remains unknown.
    \item We have not fully explored the potential of equipping our \model{} with more sophisticated agent mechanisms, such as Reflexion~\citep{shinn2023reflexion} and ReWOO~\citep{xu2023rewoo}. Further investigation into these mechanisms could yield valuable insights.
    \item This study primarily focuses on improving the agent's performance via supervised fine-tuning on expert trajectories. How to exploit exploration-based methods~\citep{song2024trial,xiong2024watch} to further optimize the agents is left for future investigation.
    \item This work is centered around building strong ReAct-style single-agent models. However, multi-agent collaboration framework has demonstrated impressive performance in handling realistic tasks. The development of strong generalized multi-agent systems based on open-source LLMs is still an under-explored area.
\end{itemize}


\section*{Acknowledgement}
We thank the anonymous reviewers for their helpful comments on this paper. This work was partially supported by National Natural Science Foundation of China (No. 62476010). 

\bibliography{custom}

\begin{thebibliography}{65}
\expandafter\ifx\csname natexlab\endcsname\relax\def\natexlab#1{#1}\fi

\bibitem[{Amini et~al.(2019)Amini, Gabriel, Lin, Koncel-Kedziorski, Choi, and Hajishirzi}]{amini2019mathqa}
Aida Amini, Saadia Gabriel, Peter Lin, Rik Koncel-Kedziorski, Yejin Choi, and Hannaneh Hajishirzi. 2019.
\newblock Mathqa: Towards interpretable math word problem solving with operation-based formalisms.
\newblock \emph{arXiv preprint arXiv:1905.13319}.

\bibitem[{Austin et~al.(2021)Austin, Odena, Nye, Bosma, Michalewski, Dohan, Jiang, Cai, Terry, Le et~al.}]{austin2021program}
Jacob Austin, Augustus Odena, Maxwell Nye, Maarten Bosma, Henryk Michalewski, David Dohan, Ellen Jiang, Carrie Cai, Michael Terry, Quoc Le, et~al. 2021.
\newblock Program synthesis with large language models.
\newblock \emph{arXiv preprint arXiv:2108.07732}.

\bibitem[{Brooks(1991)}]{brooks1991intelligence}
Rodney~A Brooks. 1991.
\newblock Intelligence without representation.
\newblock \emph{Artificial intelligence}, 47(1-3):139--159.

\bibitem[{Chen et~al.(2023{\natexlab{a}})Chen, Shu, Shareghi, Collier, Narasimhan, and Yao}]{chen2023fireact}
Baian Chen, Chang Shu, Ehsan Shareghi, Nigel Collier, Karthik Narasimhan, and Shunyu Yao. 2023{\natexlab{a}}.
\newblock Fireact: Toward language agent fine-tuning.
\newblock \emph{arXiv preprint arXiv:2310.05915}.

\bibitem[{Chen et~al.(2021)Chen, Tworek, Jun, Yuan, Pinto, Kaplan, Edwards, Burda, Joseph, Brockman et~al.}]{chen2021evaluating}
Mark Chen, Jerry Tworek, Heewoo Jun, Qiming Yuan, Henrique Ponde de~Oliveira Pinto, Jared Kaplan, Harri Edwards, Yuri Burda, Nicholas Joseph, Greg Brockman, et~al. 2021.
\newblock Evaluating large language models trained on code.
\newblock \emph{arXiv preprint arXiv:2107.03374}.

\bibitem[{Chen et~al.(2023{\natexlab{b}})Chen, Yin, Ku, Lu, Wan, Ma, Xu, Wang, and Xia}]{chen2023theoremqa}
Wenhu Chen, Ming Yin, Max Ku, Pan Lu, Yixin Wan, Xueguang Ma, Jianyu Xu, Xinyi Wang, and Tony Xia. 2023{\natexlab{b}}.
\newblock Theoremqa: A theorem-driven question answering dataset.
\newblock In \emph{The 2023 Conference on Empirical Methods in Natural Language Processing}.

\bibitem[{Chen et~al.(2024)Chen, Liu, Wang, Zhang, Liu, Lin, Chen, and Zhao}]{chen2024agent}
Zehui Chen, Kuikun Liu, Qiuchen Wang, Wenwei Zhang, Jiangning Liu, Dahua Lin, Kai Chen, and Feng Zhao. 2024.
\newblock Agent-flan: Designing data and methods of effective agent tuning for large language models.
\newblock \emph{arXiv preprint arXiv:2403.12881}.

\bibitem[{Chiang et~al.(2023)Chiang, Li, Lin, Sheng, Wu, Zhang, Zheng, Zhuang, Zhuang, Gonzalez, Stoica, and Xing}]{vicuna2023}
Wei-Lin Chiang, Zhuohan Li, Zi~Lin, Ying Sheng, Zhanghao Wu, Hao Zhang, Lianmin Zheng, Siyuan Zhuang, Yonghao Zhuang, Joseph~E. Gonzalez, Ion Stoica, and Eric~P. Xing. 2023.
\newblock Vicuna: An open-source chatbot impressing gpt-4 with 90\%* chatgpt quality.

\bibitem[{Chung et~al.(2022)Chung, Hou, Longpre, Zoph, Tay, Fedus, Li, Wang, Dehghani, Brahma et~al.}]{chung2022scaling}
Hyung~Won Chung, Le~Hou, Shayne Longpre, Barret Zoph, Yi~Tay, William Fedus, Yunxuan Li, Xuezhi Wang, Mostafa Dehghani, Siddhartha Brahma, et~al. 2022.
\newblock Scaling instruction-finetuned language models.
\newblock \emph{arXiv preprint arXiv:2210.11416}.

\bibitem[{Cobbe et~al.(2021)Cobbe, Kosaraju, Bavarian, Chen, Jun, Kaiser, Plappert, Tworek, Hilton, Nakano et~al.}]{cobbe2021training}
Karl Cobbe, Vineet Kosaraju, Mohammad Bavarian, Mark Chen, Heewoo Jun, Lukasz Kaiser, Matthias Plappert, Jerry Tworek, Jacob Hilton, Reiichiro Nakano, et~al. 2021.
\newblock Training verifiers to solve math word problems.
\newblock \emph{arXiv preprint arXiv:2110.14168}.

\bibitem[{Deng et~al.(2023)Deng, Gu, Zheng, Chen, Stevens, Wang, Sun, and Su}]{deng2023mind2web}
Xiang Deng, Yu~Gu, Boyuan Zheng, Shijie Chen, Samuel Stevens, Boshi Wang, Huan Sun, and Yu~Su. 2023.
\newblock Mind2web: Towards a generalist agent for the web.
\newblock \emph{arXiv preprint arXiv:2306.06070}.

\bibitem[{Geva et~al.(2021)Geva, Khashabi, Segal, Khot, Roth, and Berant}]{geva2021did}
Mor Geva, Daniel Khashabi, Elad Segal, Tushar Khot, Dan Roth, and Jonathan Berant. 2021.
\newblock Did aristotle use a laptop? a question answering benchmark with implicit reasoning strategies.
\newblock \emph{Transactions of the Association for Computational Linguistics}, 9:346--361.

\bibitem[{Gordon et~al.(2018)Gordon, Kembhavi, Rastegari, Redmon, Fox, and Farhadi}]{gordon2018iqa}
Daniel Gordon, Aniruddha Kembhavi, Mohammad Rastegari, Joseph Redmon, Dieter Fox, and Ali Farhadi. 2018.
\newblock Iqa: Visual question answering in interactive environments.
\newblock In \emph{Proceedings of the IEEE conference on computer vision and pattern recognition}, pages 4089--4098.

\bibitem[{Hendrycks et~al.(2020)Hendrycks, Burns, Basart, Zou, Mazeika, Song, and Steinhardt}]{hendrycks2020measuring}
Dan Hendrycks, Collin Burns, Steven Basart, Andy Zou, Mantas Mazeika, Dawn Song, and Jacob Steinhardt. 2020.
\newblock Measuring massive multitask language understanding.
\newblock \emph{arXiv preprint arXiv:2009.03300}.

\bibitem[{Hendrycks et~al.(2021)Hendrycks, Burns, Kadavath, Arora, Basart, Tang, Song, and Steinhardt}]{hendrycks2021measuring}
Dan Hendrycks, Collin Burns, Saurav Kadavath, Akul Arora, Steven Basart, Eric Tang, Dawn Song, and Jacob Steinhardt. 2021.
\newblock Measuring mathematical problem solving with the math dataset.
\newblock \emph{arXiv preprint arXiv:2103.03874}.

\bibitem[{Hussein et~al.(2017)Hussein, Gaber, Elyan, and Jayne}]{hussein2017imitation}
Ahmed Hussein, Mohamed~Medhat Gaber, Eyad Elyan, and Chrisina Jayne. 2017.
\newblock Imitation learning: A survey of learning methods.
\newblock \emph{ACM Computing Surveys (CSUR)}, 50(2):1--35.

\bibitem[{Jennings et~al.(1998)Jennings, Sycara, and Wooldridge}]{jennings1998roadmap}
Nicholas~R Jennings, Katia Sycara, and Michael Wooldridge. 1998.
\newblock A roadmap of agent research and development.
\newblock \emph{Autonomous agents and multi-agent systems}, 1:7--38.

\bibitem[{Jiang et~al.(2023)Jiang, Sablayrolles, Mensch, Bamford, Chaplot, Casas, Bressand, Lengyel, Lample, Saulnier et~al.}]{jiang2023mistral}
Albert~Q Jiang, Alexandre Sablayrolles, Arthur Mensch, Chris Bamford, Devendra~Singh Chaplot, Diego de~las Casas, Florian Bressand, Gianna Lengyel, Guillaume Lample, Lucile Saulnier, et~al. 2023.
\newblock Mistral 7b.
\newblock \emph{arXiv preprint arXiv:2310.06825}.

\bibitem[{Joshi et~al.(2017)Joshi, Choi, Weld, and Zettlemoyer}]{joshi2017triviaqa}
Mandar Joshi, Eunsol Choi, Daniel~S Weld, and Luke Zettlemoyer. 2017.
\newblock Triviaqa: A large scale distantly supervised challenge dataset for reading comprehension.
\newblock \emph{arXiv preprint arXiv:1705.03551}.

\bibitem[{Kim et~al.(2023)Kim, Baldi, and McAleer}]{kim2023language}
Geunwoo Kim, Pierre Baldi, and Stephen McAleer. 2023.
\newblock Language models can solve computer tasks.
\newblock \emph{arXiv preprint arXiv:2303.17491}.

\bibitem[{Kojima et~al.(2022)Kojima, Gu, Reid, Matsuo, and Iwasawa}]{kojima2022large}
Takeshi Kojima, Shixiang~Shane Gu, Machel Reid, Yutaka Matsuo, and Yusuke Iwasawa. 2022.
\newblock Large language models are zero-shot reasoners.
\newblock \emph{Advances in neural information processing systems}, 35:22199--22213.

\bibitem[{Lee et~al.(2023)Lee, Hunter, and Ruiz}]{lee2023platypus}
Ariel~N Lee, Cole~J Hunter, and Nataniel Ruiz. 2023.
\newblock Platypus: Quick, cheap, and powerful refinement of llms.
\newblock \emph{arXiv preprint arXiv:2308.07317}.

\bibitem[{Li et~al.(2023)Li, Zhang, Dubois, Taori, Gulrajani, Guestrin, Liang, and Hashimoto}]{alpaca_eval}
Xuechen Li, Tianyi Zhang, Yann Dubois, Rohan Taori, Ishaan Gulrajani, Carlos Guestrin, Percy Liang, and Tatsunori~B. Hashimoto. 2023.
\newblock Alpacaeval: An automatic evaluator of instruction-following models.
\newblock \url{https://github.com/tatsu-lab/alpaca_eval}.

\bibitem[{Liang et~al.(2022)Liang, Bommasani, Lee, Tsipras, Soylu, Yasunaga, Zhang, Narayanan, Wu, Kumar et~al.}]{liang2022holistic}
Percy Liang, Rishi Bommasani, Tony Lee, Dimitris Tsipras, Dilara Soylu, Michihiro Yasunaga, Yian Zhang, Deepak Narayanan, Yuhuai Wu, Ananya Kumar, et~al. 2022.
\newblock Holistic evaluation of language models.
\newblock \emph{arXiv preprint arXiv:2211.09110}.

\bibitem[{Liu et~al.(2023)Liu, Yu, Zhang, Xu, Lei, Lai, Gu, Ding, Men, Yang et~al.}]{liu2023agentbench}
Xiao Liu, Hao Yu, Hanchen Zhang, Yifan Xu, Xuanyu Lei, Hanyu Lai, Yu~Gu, Hangliang Ding, Kaiwen Men, Kejuan Yang, et~al. 2023.
\newblock Agentbench: Evaluating llms as agents.
\newblock \emph{arXiv preprint arXiv:2308.03688}.

\bibitem[{Luo et~al.(2023)Luo, Xu, Zhao, Sun, Geng, Hu, Tao, Ma, Lin, and Jiang}]{luo2023wizardcoder}
Ziyang Luo, Can Xu, Pu~Zhao, Qingfeng Sun, Xiubo Geng, Wenxiang Hu, Chongyang Tao, Jing Ma, Qingwei Lin, and Daxin Jiang. 2023.
\newblock Wizardcoder: Empowering code large language models with evol-instruct.

\bibitem[{Meta(2024)}]{llama3}
Meta. 2024.
\newblock \href {https://ai.meta.com/blog/meta-llama-3/} {Introducing meta llama 3: The most capable openly available llm to date}.

\bibitem[{Mukherjee et~al.(2023)Mukherjee, Mitra, Jawahar, Agarwal, Palangi, and Awadallah}]{mukherjee2023orca}
Subhabrata Mukherjee, Arindam Mitra, Ganesh Jawahar, Sahaj Agarwal, Hamid Palangi, and Ahmed Awadallah. 2023.
\newblock Orca: Progressive learning from complex explanation traces of gpt-4.
\newblock \emph{arXiv preprint arXiv:2306.02707}.

\bibitem[{OpenAI(2022)}]{chatgpt}
OpenAI. 2022.
\newblock Introducing chatgpt.

\bibitem[{OpenAI(2023)}]{gpt4}
OpenAI. 2023.
\newblock Gpt-4 technical report.
\newblock \emph{arXiv}, pages 2303--08774.

\bibitem[{Paszke et~al.(2019)Paszke, Gross, Massa, Lerer, Bradbury, Chanan, Killeen, Lin, Gimelshein, Antiga et~al.}]{paszke2019pytorch}
Adam Paszke, Sam Gross, Francisco Massa, Adam Lerer, James Bradbury, Gregory Chanan, Trevor Killeen, Zeming Lin, Natalia Gimelshein, Luca Antiga, et~al. 2019.
\newblock Pytorch: An imperative style, high-performance deep learning library.
\newblock \emph{Advances in neural information processing systems}, 32.

\bibitem[{Press et~al.(2022)Press, Zhang, Min, Schmidt, Smith, and Lewis}]{press2022measuring}
Ofir Press, Muru Zhang, Sewon Min, Ludwig Schmidt, Noah~A Smith, and Mike Lewis. 2022.
\newblock Measuring and narrowing the compositionality gap in language models.
\newblock \emph{arXiv preprint arXiv:2210.03350}.

\bibitem[{Richards(2023)}]{autogpt}
Toran~Bruce Richards. 2023.
\newblock Auto-gpt: An autonomous gpt-4 experiment.

\bibitem[{Roziere et~al.(2023)Roziere, Gehring, Gloeckle, Sootla, Gat, Tan, Adi, Liu, Remez, Rapin et~al.}]{roziere2023code}
Baptiste Roziere, Jonas Gehring, Fabian Gloeckle, Sten Sootla, Itai Gat, Xiaoqing~Ellen Tan, Yossi Adi, Jingyu Liu, Tal Remez, J{\'e}r{\'e}my Rapin, et~al. 2023.
\newblock Code llama: Open foundation models for code.
\newblock \emph{arXiv preprint arXiv:2308.12950}.

\bibitem[{Sanh et~al.(2021)Sanh, Webson, Raffel, Bach, Sutawika, Alyafeai, Chaffin, Stiegler, Scao, Raja et~al.}]{sanh2021multitask}
Victor Sanh, Albert Webson, Colin Raffel, Stephen~H Bach, Lintang Sutawika, Zaid Alyafeai, Antoine Chaffin, Arnaud Stiegler, Teven~Le Scao, Arun Raja, et~al. 2021.
\newblock Multitask prompted training enables zero-shot task generalization.
\newblock \emph{arXiv preprint arXiv:2110.08207}.

\bibitem[{Shinn et~al.(2023)Shinn, Labash, and Gopinath}]{shinn2023reflexion}
Noah Shinn, Beck Labash, and Ashwin Gopinath. 2023.
\newblock Reflexion: an autonomous agent with dynamic memory and self-reflection.
\newblock \emph{arXiv preprint arXiv:2303.11366}.

\bibitem[{Shridhar et~al.(2020{\natexlab{a}})Shridhar, Thomason, Gordon, Bisk, Han, Mottaghi, Zettlemoyer, and Fox}]{shridhar2020alfred}
Mohit Shridhar, Jesse Thomason, Daniel Gordon, Yonatan Bisk, Winson Han, Roozbeh Mottaghi, Luke Zettlemoyer, and Dieter Fox. 2020{\natexlab{a}}.
\newblock Alfred: A benchmark for interpreting grounded instructions for everyday tasks.
\newblock In \emph{Proceedings of the IEEE/CVF conference on computer vision and pattern recognition}, pages 10740--10749.

\bibitem[{Shridhar et~al.(2020{\natexlab{b}})Shridhar, Yuan, C{\^o}t{\'e}, Bisk, Trischler, and Hausknecht}]{shridhar2020alfworld}
Mohit Shridhar, Xingdi Yuan, Marc-Alexandre C{\^o}t{\'e}, Yonatan Bisk, Adam Trischler, and Matthew Hausknecht. 2020{\natexlab{b}}.
\newblock Alfworld: Aligning text and embodied environments for interactive learning.
\newblock \emph{arXiv preprint arXiv:2010.03768}.

\bibitem[{Song et~al.(2023)Song, Xiong, Zhu, Li, Wang, Tian, and Li}]{song2023restgpt}
Yifan Song, Weimin Xiong, Dawei Zhu, Cheng Li, Ke~Wang, Ye~Tian, and Sujian Li. 2023.
\newblock Restgpt: Connecting large language models with real-world applications via restful apis.
\newblock \emph{arXiv preprint arXiv:2306.06624}.

\bibitem[{Song et~al.(2024)Song, Yin, Yue, Huang, Li, and Lin}]{song2024trial}
Yifan Song, Da~Yin, Xiang Yue, Jie Huang, Sujian Li, and Bill~Yuchen Lin. 2024.
\newblock Trial and error: Exploration-based trajectory optimization for llm agents.
\newblock \emph{arXiv preprint arXiv:2403.02502}.

\bibitem[{Taori et~al.(2023)Taori, Gulrajani, Zhang, Dubois, Li, Guestrin, Liang, and Hashimoto}]{alpaca}
Rohan Taori, Ishaan Gulrajani, Tianyi Zhang, Yann Dubois, Xuechen Li, Carlos Guestrin, Percy Liang, and Tatsunori~B. Hashimoto. 2023.
\newblock Stanford alpaca: An instruction-following llama model.
\newblock \url{https://github.com/tatsu-lab/stanford_alpaca}.

\bibitem[{Touvron et~al.(2023)Touvron, Martin, Stone, Albert, Almahairi, Babaei, Bashlykov, Batra, Bhargava, Bhosale et~al.}]{touvron2023llama}
Hugo Touvron, Louis Martin, Kevin Stone, Peter Albert, Amjad Almahairi, Yasmine Babaei, Nikolay Bashlykov, Soumya Batra, Prajjwal Bhargava, Shruti Bhosale, et~al. 2023.
\newblock Llama 2: Open foundation and fine-tuned chat models.
\newblock \emph{arXiv preprint arXiv:2307.09288}.

\bibitem[{Wang et~al.(2022{\natexlab{a}})Wang, Jansen, C{\^o}t{\'e}, and Ammanabrolu}]{wang2022scienceworld}
Ruoyao Wang, Peter Jansen, Marc-Alexandre C{\^o}t{\'e}, and Prithviraj Ammanabrolu. 2022{\natexlab{a}}.
\newblock Scienceworld: Is your agent smarter than a 5th grader?
\newblock \emph{arXiv preprint arXiv:2203.07540}.

\bibitem[{Wang et~al.(2023)Wang, Wang, Liu, Chen, Yuan, Peng, and Ji}]{wang2023mint}
Xingyao Wang, Zihan Wang, Jiateng Liu, Yangyi Chen, Lifan Yuan, Hao Peng, and Heng Ji. 2023.
\newblock Mint: Evaluating llms in multi-turn interaction with tools and language feedback.
\newblock \emph{arXiv preprint arXiv:2309.10691}.

\bibitem[{Wang et~al.(2022{\natexlab{b}})Wang, Mishra, Alipoormolabashi, Kordi, Mirzaei, Arunkumar, Ashok, Dhanasekaran, Naik, Stap et~al.}]{wang2022super}
Yizhong Wang, Swaroop Mishra, Pegah Alipoormolabashi, Yeganeh Kordi, Amirreza Mirzaei, Anjana Arunkumar, Arjun Ashok, Arut~Selvan Dhanasekaran, Atharva Naik, David Stap, et~al. 2022{\natexlab{b}}.
\newblock Super-naturalinstructions: Generalization via declarative instructions on 1600+ nlp tasks.
\newblock \emph{arXiv preprint arXiv:2204.07705}.

\bibitem[{Wei et~al.(2021)Wei, Bosma, Zhao, Guu, Yu, Lester, Du, Dai, and Le}]{wei2021finetuned}
Jason Wei, Maarten Bosma, Vincent~Y Zhao, Kelvin Guu, Adams~Wei Yu, Brian Lester, Nan Du, Andrew~M Dai, and Quoc~V Le. 2021.
\newblock Finetuned language models are zero-shot learners.
\newblock \emph{arXiv preprint arXiv:2109.01652}.

\bibitem[{Wei et~al.(2022)Wei, Wang, Schuurmans, Bosma, Xia, Chi, Le, Zhou et~al.}]{wei2022chain}
Jason Wei, Xuezhi Wang, Dale Schuurmans, Maarten Bosma, Fei Xia, Ed~Chi, Quoc~V Le, Denny Zhou, et~al. 2022.
\newblock Chain-of-thought prompting elicits reasoning in large language models.
\newblock \emph{Advances in Neural Information Processing Systems}, 35:24824--24837.

\bibitem[{Weihs et~al.(2021)Weihs, Deitke, Kembhavi, and Mottaghi}]{RoomR}
Luca Weihs, Matt Deitke, Aniruddha Kembhavi, and Roozbeh Mottaghi. 2021.
\newblock Visual room rearrangement.
\newblock In \emph{IEEE/CVF Conference on Computer Vision and Pattern Recognition (CVPR)}.

\bibitem[{Xiong et~al.(2024)Xiong, Song, Zhao, Wu, Wang, Wang, Li, Peng, and Li}]{xiong2024watch}
Weimin Xiong, Yifan Song, Xiutian Zhao, Wenhao Wu, Xun Wang, Ke~Wang, Cheng Li, Wei Peng, and Sujian Li. 2024.
\newblock Watch every step! llm agent learning via iterative step-level process refinement.
\newblock \emph{arXiv preprint arXiv:2406.11176}.

\bibitem[{Xu et~al.(2023{\natexlab{a}})Xu, Peng, Lei, Mukherjee, Liu, and Xu}]{xu2023rewoo}
Binfeng Xu, Zhiyuan Peng, Bowen Lei, Subhabrata Mukherjee, Yuchen Liu, and Dongkuan Xu. 2023{\natexlab{a}}.
\newblock Rewoo: Decoupling reasoning from observations for efficient augmented language models.
\newblock \emph{arXiv preprint arXiv:2305.18323}.

\bibitem[{Xu et~al.(2023{\natexlab{b}})Xu, Su, Xing, Mi, Liu, Shi, Hui, Zhou, Liu, Xie et~al.}]{xu2023lemur}
Yiheng Xu, Hongjin Su, Chen Xing, Boyu Mi, Qian Liu, Weijia Shi, Binyuan Hui, Fan Zhou, Yitao Liu, Tianbao Xie, et~al. 2023{\natexlab{b}}.
\newblock Lemur: Harmonizing natural language and code for language agents.
\newblock \emph{arXiv preprint arXiv:2310.06830}.

\bibitem[{Yang et~al.(2023)Yang, Prabhakar, Narasimhan, and Yao}]{yang2023intercode}
John Yang, Akshara Prabhakar, Karthik Narasimhan, and Shunyu Yao. 2023.
\newblock Intercode: Standardizing and benchmarking interactive coding with execution feedback.
\newblock \emph{arXiv preprint arXiv:2306.14898}.

\bibitem[{Yang et~al.(2024)Yang, Liu, Wu, Yang, Fung, Li, Huang, Cao, Wang, Wang et~al.}]{yang2024if}
Ke~Yang, Jiateng Liu, John Wu, Chaoqi Yang, Yi~R Fung, Sha Li, Zixuan Huang, Xu~Cao, Xingyao Wang, Yiquan Wang, et~al. 2024.
\newblock If llm is the wizard, then code is the wand: A survey on how code empowers large language models to serve as intelligent agents.
\newblock \emph{arXiv preprint arXiv:2401.00812}.

\bibitem[{Yang et~al.(2018)Yang, Qi, Zhang, Bengio, Cohen, Salakhutdinov, and Manning}]{yang2018hotpotqa}
Zhilin Yang, Peng Qi, Saizheng Zhang, Yoshua Bengio, William~W Cohen, Ruslan Salakhutdinov, and Christopher~D Manning. 2018.
\newblock Hotpotqa: A dataset for diverse, explainable multi-hop question answering.
\newblock \emph{arXiv preprint arXiv:1809.09600}.

\bibitem[{Yao et~al.(2022{\natexlab{a}})Yao, Chen, Yang, and Narasimhan}]{yao2022webshop}
Shunyu Yao, Howard Chen, John Yang, and Karthik Narasimhan. 2022{\natexlab{a}}.
\newblock Webshop: Towards scalable real-world web interaction with grounded language agents.
\newblock \emph{Advances in Neural Information Processing Systems}, 35:20744--20757.

\bibitem[{Yao et~al.(2022{\natexlab{b}})Yao, Zhao, Yu, Du, Shafran, Narasimhan, and Cao}]{yao2022react}
Shunyu Yao, Jeffrey Zhao, Dian Yu, Nan Du, Izhak Shafran, Karthik Narasimhan, and Yuan Cao. 2022{\natexlab{b}}.
\newblock React: Synergizing reasoning and acting in language models.
\newblock \emph{arXiv preprint arXiv:2210.03629}.

\bibitem[{Yin et~al.(2023)Yin, Brahman, Ravichander, Chandu, Chang, Choi, and Lin}]{yin2023lumos}
Da~Yin, Faeze Brahman, Abhilasha Ravichander, Khyathi Chandu, Kai-Wei Chang, Yejin Choi, and Bill~Yuchen Lin. 2023.
\newblock Lumos: Learning agents with unified data, modular design, and open-source llms.
\newblock \emph{arXiv preprint arXiv:2311.05657}.

\bibitem[{Yu et~al.(2023)Yu, Jiang, Shi, Yu, Liu, Zhang, Kwok, Li, Weller, and Liu}]{yu2023metamath}
Longhui Yu, Weisen Jiang, Han Shi, Jincheng Yu, Zhengying Liu, Yu~Zhang, James~T Kwok, Zhenguo Li, Adrian Weller, and Weiyang Liu. 2023.
\newblock Metamath: Bootstrap your own mathematical questions for large language models.
\newblock \emph{arXiv preprint arXiv:2309.12284}.

\bibitem[{Yue et~al.(2023)Yue, Qu, Zhang, Fu, Huang, Sun, Su, and Chen}]{yue2023mammoth}
Xiang Yue, Xingwei Qu, Ge~Zhang, Yao Fu, Wenhao Huang, Huan Sun, Yu~Su, and Wenhu Chen. 2023.
\newblock Mammoth: Building math generalist models through hybrid instruction tuning.
\newblock \emph{arXiv preprint arXiv:2309.05653}.

\bibitem[{Zeng et~al.(2023)Zeng, Liu, Lu, Wang, Liu, Dong, and Tang}]{zeng2023agenttuning}
Aohan Zeng, Mingdao Liu, Rui Lu, Bowen Wang, Xiao Liu, Yuxiao Dong, and Jie Tang. 2023.
\newblock Agenttuning: Enabling generalized agent abilities for llms.
\newblock \emph{arXiv preprint arXiv:2310.12823}.

\bibitem[{Zhang et~al.(2024)Zhang, Lan, Murthy, Liu, Yao, Tan, Hoang, Yang, Feng, Liu et~al.}]{zhang2024agentohana}
Jianguo Zhang, Tian Lan, Rithesh Murthy, Zhiwei Liu, Weiran Yao, Juntao Tan, Thai Hoang, Liangwei Yang, Yihao Feng, Zuxin Liu, et~al. 2024.
\newblock Agentohana: Design unified data and training pipeline for effective agent learning.
\newblock \emph{arXiv preprint arXiv:2402.15506}.

\bibitem[{Zhang et~al.(2023)Zhang, Dong, Li, Zhang, Sun, Wang, Li, Hu, Zhang, Wu et~al.}]{zhang2023instruction}
Shengyu Zhang, Linfeng Dong, Xiaoya Li, Sen Zhang, Xiaofei Sun, Shuhe Wang, Jiwei Li, Runyi Hu, Tianwei Zhang, Fei Wu, et~al. 2023.
\newblock Instruction tuning for large language models: A survey.
\newblock \emph{arXiv preprint arXiv:2308.10792}.

\bibitem[{Zheng et~al.(2023{\natexlab{a}})Zheng, Chiang, Sheng, Zhuang, Wu, Zhuang, Lin, Li, Li, Xing, Zhang, Gonzalez, and Stoica}]{zheng2023judging}
Lianmin Zheng, Wei-Lin Chiang, Ying Sheng, Siyuan Zhuang, Zhanghao Wu, Yonghao Zhuang, Zi~Lin, Zhuohan Li, Dacheng Li, Eric.~P Xing, Hao Zhang, Joseph~E. Gonzalez, and Ion Stoica. 2023{\natexlab{a}}.
\newblock \href {http://arxiv.org/abs/2306.05685} {Judging llm-as-a-judge with mt-bench and chatbot arena}.

\bibitem[{Zheng et~al.(2023{\natexlab{b}})Zheng, Wang, Jia, and Tong}]{langsuite2023}
Zilong Zheng, Mengmeng Wang, Zixia Jia, and Baichen Tong. 2023{\natexlab{b}}.
\newblock Langsuite: Controlling, planning, and interacting with large language models in embodied text environments.

\bibitem[{Zhou et~al.(2023)Zhou, Xu, Zhu, Zhou, Lo, Sridhar, Cheng, Bisk, Fried, Alon et~al.}]{zhou2023webarena}
Shuyan Zhou, Frank~F Xu, Hao Zhu, Xuhui Zhou, Robert Lo, Abishek Sridhar, Xianyi Cheng, Yonatan Bisk, Daniel Fried, Uri Alon, et~al. 2023.
\newblock Webarena: A realistic web environment for building autonomous agents.
\newblock \emph{arXiv preprint arXiv:2307.13854}.

\end{thebibliography}

\appendix

\clearpage

\section{Details of Tasks in \dataset{}}
\label{app:dataset}

\paragraph{Reasoning Tasks}

HotpotQA~\citep{yang2018hotpotqa} is a question answering dataset featuring multi-hop reasoning.
StrategyQA~\citep{geva2021did} is another question answering task where the required reasoning steps are implicit in the question and should be inferred using a strategy.
TriviaQA~\citep{joshi2017triviaqa} is a dataset consisting of complex compositional questions that require multi-evidence reasoning.
In our work, we repurpose these three datasets to interaction environments by incorporating a search engine tool.
We employ the GPT-exploration pipeline and filter out failed cases to build the gold trajectories.

For our held-out evaluation, we use Bamboogle~\citep{press2022measuring}, which is made up of questions that need compositional reasoning and are unable to be directly answered by Google.

\paragraph{Math Tasks}

GSM8K~\citep{cobbe2021training} is a dataset of diverse grade school math problems created by humans. 
Each problem in GSM8K comes with an official solution path. 
In our work, we leverage the power of GPT-3.5-Turbo to transform these solution paths into interaction trajectories.

MathQA~\citep{amini2019mathqa} is a large-scale multiple-choice math problem dataset covering multiple math domains.
MATH~\citep{press2022measuring} contains challenging mathematics problems from high school math competitions.
To adapt these two datasets into interaction environments, we employ a Python interpreter and employ the GPT-exploration pipeline to construct the trajectories.

For the held-out task, we use TheoremQA~\citep{chen2023theoremqa}, a theorem-driven question answering dataset composing of high-quality questions from math, physics, EE\&CS, and finance.
We implement Python interpreter and Wikipedia tools to construct the corresponding interactive environment.

\paragraph{Programming Tasks}

InterCode~\citep{yang2023intercode} is a benchmark for evaluating language models on interactive programming tasks. 
In this task, agents are required to respond to natural language requests by interacting with a software system, such as a database or terminal.
Our work focuses on evaluating the programming ability of agents using two environments: IC-Bash and IC-SQL. 
IC-Bash is specifically used for the held-out evaluation of agents.

APPS~\citep{hendrycks2021measuring} is a benchmark focused on Python code generation, encompassing a range of difficulty levels from introductory to competition level.
We utilize GPT-3.5-Turbo to reformat the instances in this dataset and construct the trajectories.

HumanEval~\citep{chen2021evaluating} is a dataset designed to measure functional correctness for synthesizing programs from docstrings.
MBPP~\citep{austin2021program} consists of around 1,000 crowd-sourced Python programming problems.
For both of these datasets, we employ the GPT-exploration pipeline to annotate the interaction trajectories. 
Subsequently, we employ the answer forcing method to re-annotate the cases where GPT failed.

\paragraph{Web Tasks}
Mind2Web~\citep{deng2023mind2web} is a dataset for developing and evaluating generalist agents for the web that can follow language instructions to complete complex tasks on any website.
WebArena~\citep{zhou2023webarena} builds realistic web environments for agents to execute tasks.
Even GPT-4 struggles with these tasks, so we utilize a teacher forcing and break down the complete interaction trajectory into multiple single steps.
Then GPT-3.5-Turbo is employed to annotate the rationales.

WebShop~\citep{yao2022webshop} is a simulated e-commerce website environment with real-world products and crowd-sourced text instructions. 
For 1571 official human annotated trajectories, we employ GPT-3.5-Turbo to reformat them and annotate rationales.
Additionally, we incorporate trajectories generated through GPT-exploration, which have final rewards exceeding 0.3.

For our held-out task, we utilize MiniWoB++~\citep{kim2023language}, a diverse collection of over 100 web interaction environments, to formulate our benchmark.

\begin{table*}[t]
\centering
\resizebox{0.9\linewidth}{!}{
\begin{tabular}{llccccc}
\toprule
Dataset & Model & $R_\mathrm{train}$ & $R_\mathrm{pseudo}$ & $R_\mathrm{test}$ & $\Delta_1$ & $\Delta_2$ \\
\midrule
\multirow{2}{*}{AgentInstruct~\citep{zeng2023agenttuning}} & Llama-2-7B-Chat & 17.8 & 17.5 & 15.8 & -0.3 & -2.0 \\
& +$\mathcal{D}_\mathrm{train}$ & 72.5 & 72.6 & 62.4 & +0.1 & -10.1 \\
\midrule
\multirow{2}{*}{\dataset{} (Ours)} & Llama-2-7B-Chat & 16.2 & 16.5 & 16.0 & +0.3 & -0.2 \\
& +$\mathcal{D}_\mathrm{train}$ & 73.3 & 62.3 & 62.8 & -11.0 & -10.5 \\
\bottomrule
\end{tabular}
}
\caption{
The average reward of WebShop on different instruction sets. 
We compare the reward $R_\mathrm{train}$, $R_\mathrm{pseudo}$, $R_\mathrm{test}$ on the training set $\mathcal{D}_\mathrm{train}$, a pseudo test set held-out from the original training set $\mathcal{D}_\mathrm{pseudo}$, and original test set $\mathcal{D}_\mathrm{test}$ respectively. 
We also reports two key metrics: $\Delta_1=R_\mathrm{pseudo}-R_\mathrm{train}$ and $\Delta_2=R_\mathrm{test}-R_\mathrm{train}$, as the indicators of the difficulty differences between datasets.
}
\label{tab:bias}
\end{table*}

\paragraph{Embodied AI Tasks}

ALFWorld~\citep{shridhar2020alfworld} contains interactive TextWorld environments that parallel embodied worlds in the ALFRED dataset~\citep{shridhar2020alfred}.
This dataset provides human-annotated gold trajectories for imitation learning.
RoomR~\citep{RoomR} is an embodied AI dataset which requires agents to restore the initial configurations of all objects within a room.
IQA~\citep{gordon2018iqa} is a question answering task that requires an agent to interact with a dynamic visual environment.
In our work, we utilize the text versions of RoomR and IQA developed by \citet{langsuite2023}.
We employ a depth-first-search algorithm to build the gold action sequences for RoomR and IQA.
We then leverage GPT-3.5-Turbo to annotate the corresponding rationales.

For the held-out evaluation, we utilize ScienceWorld~\citep{wang2022scienceworld}, a text-based virtual environment which encompasses various elementary science experiment tasks, including thermodynamics and electrical circuits.

\section{Difficulty Bias in Trajectory Collection}
\label{app:bias}

In this section, we conduct a experiment to verify the existence of difficulty bias introduced by the trajectory annotation pipeline widely used in recent studies~\citep{chen2023fireact,zeng2023agenttuning}.
Specifically, we choose WebShop trajectories in \dataset{} and AgentInstruct~\citep{zeng2023agenttuning} to conduct the experiment.
For AgentInstruct and \dataset{}, we select 300 instances as the training set $\mathcal{D}_\mathrm{train}$, 50 instances as the pseudo test set $\mathcal{D}_\mathrm{pseudo}$.
We also include the original WebShop test set $\mathcal{D}_\mathrm{test}$.

For a dataset conforming to the \textit{i.i.d.} assumption, the instances in $\mathcal{D}_\mathrm{train}$, $\mathcal{D}_\mathrm{pseudo}$, $\mathcal{D}_\mathrm{test}$ are sampled from the same distribution.
Therefore, the expected behavior is that the evaluation results on $\mathcal{D}_\mathrm{pseudo}$ and $\mathcal{D}_\mathrm{test}$ should be consistent.
Furthermore, an agent trained on $\mathcal{D}_\mathrm{train}$ should ideally perform better on $\mathcal{D}_\mathrm{train}$ compared to $\mathcal{D}_\mathrm{pseudo}$ and $\mathcal{D}_\mathrm{test}$.

Table \ref{tab:bias} illustrates the performance of untrained Llama-2-7B-Chat and the trained agent on different sets.
For AgentInstruct, both models exhibit worse performance on $\mathcal{D}_\mathrm{test}$ compared to $\mathcal{D}_\mathrm{pseudo}$, indicating that instances in AgentInstruct are considerably easier than those in the original test set.
Conversely, for \dataset{}, the agents have close performance on $\mathcal{D}_\mathrm{pseudo}$ and $\mathcal{D}_\mathrm{test}$, aligning with our expectations.
The agent trained on our dataset also outperforms the agent trained on AgentInstruct when evaluated on $\mathcal{D}_\mathrm{test}$.
These experiments highlight that the GPT-exploration trajectory annotation pipeline can introduce difficulty bias in the training set, potentially compromising the generalizability of trained agents.

\section{CoT Rationales Generated by Different LLMs}
\label{app:rationale_compare}

\begin{table}[t]
\centering
\resizebox{0.8\linewidth}{!}{
\begin{tabular}{lcc}
\toprule
Rationale & IC-SQL & WebShop \\
\midrule
GPT-4 & 58.5 & 63.4 \\
GPT-3.5-Turbo & 58.8 & 63.2 \\
\bottomrule
\end{tabular}
}
\caption{
Comparison of rationales generated by different LLMs.
}
\label{tab:rationale_compare}
\end{table}

Since providing explanation for gold actions is relatively easy task, we employ GPT-3.5-Turbo as the primary LLM in the rationale annotation process for \dataset{}.
Here we compare the difference of rationale generated by different LLMs.
Specifically, we select IC-SQL and WebShop to conduct the experiments.
As shown in Table~\ref{tab:rationale_compare}, agents training with rationale generated by GPT-4 and GPT-3.5-Turbo have little performance gap.

\section{Quality Control of \dataset{}}
\label{app:quality}

In Section \ref{sec:annotation}, we incorporate heuristic and GPT-based methods to construct \dataset{}, which can mitigate the difficulty bias problem in the previous annotation pipeline.
In this section, we propose to perform a human evaluation to assess the quality of \dataset{}.
To achieve this, we employ 5 human annotators who are instructed to choose the better trajectory from two anonymous candidate options.
Here, we select two representative tasks: IC-SQL to assess the quality of answer forcing annotation, and WebShop to evaluate the quality of trajectory reformatting.
For IC-SQL, we compare 100 trajectories generated by answer forcing with those generated through GPT exploration.
For WebShop, we select 80 trajectories from \dataset{} and \citet{zeng2023agenttuning} which correspond to the same task instance.

\begin{table}[t]
\centering
\resizebox{0.8\linewidth}{!}{
\begin{tabular}{lcccc}
\toprule
Dataset & Win & Lose & Tie & Total \\
\midrule
IC-SQL & 11 & 16 & 73 & 100 \\
WebShop & 12 & 10 & 58 & 80 \\
\bottomrule
\end{tabular}
}
\caption{
Human evaluation of the data quality for \dataset{}. For IC-SQL, we compare trajectories generated through answer forcing with those generated through exploration. For WebShop, we compare our constructed trajectories with the trajectories constructed by \citet{zeng2023agenttuning}.
}
\label{tab:quality}
\end{table}

As shown in Table \ref{tab:quality}, for most cases, trajectories generated by answer forcing or reformatting have the same quality as GPT exploration.
Therefore, we can conclude that our trajectory annotation process can achieve comparable quality with previous methods~\citep{chen2023fireact,zeng2023agenttuning} while mitigating the difficulty bias.

\begin{table*}[b]
\centering
\resizebox{\linewidth}{!}{
\begin{tabular}{lccccccc}
\toprule
\multirow{2}{*}{\textbf{Dataset}} & \multirow{2}{*}{\textbf{\#Inst}} & \multicolumn{2}{c}{\textbf{\dataset{}}} & \multicolumn{2}{c}{\textbf{ShareGPT}} & \multicolumn{2}{c}{\textbf{Evol-CodeAlpaca}} \\
\cmidrule(l){3-4} \cmidrule(l){5-6} \cmidrule(l){7-8} 
& & 9-Gram Rate & 13-Gram Rate & 9-Gram Rate & 13-Gram Rate & 9-Gram Rate & 13-Gram Rate \\
\midrule
\multicolumn{8}{c}{\textit{Held-in Tasks}} \\
\midrule
HotpotQA & 100 & \textcolor{red}{1\%} & 0\% & 0\% & 0\% & 0\% & 0\% \\
StrategyQA & 100 & \textcolor{red}{20\%} & \textcolor{red}{12\%} & 0\% & 0\% & 0\% & 0\% \\
GSM8K & 100 & \textcolor{red}{3\%} & 0\% & 0\% & 0\% & 0\% & 0\% \\
MATH & 100 & \textcolor{red}{15\%} & \textcolor{red}{4\%} & 0\% & 0\% & \textcolor{red}{2\%} & 0\% \\
IC-SQL & 100 & \textcolor{red}{7\%} & 0\% & 0\% & 0\% & \textcolor{red}{1\%} & 0\% \\
MBPP & 100 & \textcolor{red}{12\%} & \textcolor{red}{1\%} & \textcolor{red}{7\%} & \textcolor{red}{3\%} & \textcolor{red}{18\%} & \textcolor{red}{4\%} \\
Mind2Web & 1173 & \textcolor{red}{8\%} & \textcolor{red}{3\%} & 0\% & 0\% & 0\% & 0\% \\
WebShop & 200 & \textcolor{red}{41\%} & \textcolor{red}{14\%} & 0\% & 0\% & 0\% & 0\% \\
ALFWorld & 134 & \textcolor{red}{14\%} & \textcolor{red}{8\%} & 0\% & 0\% & 0\% & 0\% \\
\midrule
\multicolumn{8}{c}{\textit{Held-out Tasks}} \\
\midrule
Bamboogle & 126 & 0\% & 0\% & 0\% & 0\% & 0\% & 0\% \\
ThreomQA & 100 & 0\% & 0\% & 0\% & 0\% & 0\% & 0\% \\
IC-Bash & 200 & 0\% & 0\% & 0\% & 0\% & 0\% & 0\% \\
MiniWoB++ & 460 & 0\% & 0\% & 0\% & 0\% & \textcolor{red}{2\%} & 0\% \\
SciWorld & 270 & 0\% & 0\% & 0\% & 0\% & 0\% & 0\% \\
\bottomrule
\end{tabular}
}
\caption{
Data contamination analysis.
}
\label{tab:contam}
\end{table*}

\section{Data Contamination}
\label{app:contam}

When training \model{}, we construct a data mixture consisting of trajectory data (\dataset{}), generalist instruction data (ShareGPT), and code data (Evol-CodeAlpaca).
However, it is important to address the concern of potential data contamination, which could result in an overestimation of performance.
Therefore, we perform a contamination analysis by comparing our evaluation set with \dataset{}, ShareGPT, and Evol-CodeAlpaca.
Following \citet{liang2022holistic}, we heuristically match 9-grams and 13-grams from the instances in the test set with the training set data.
Table \ref{tab:contam} displays the proportion of instances which exhibit an overlap with the training data.

First, we observe a high contamination rate for held-in tasks with \dataset{}.
After manually examining these instances, we have some findings.
In the case of StrategyQA, we discovered that all instances followed a question format that could be answered with a simple "yes" or "no," potentially resulting in a high n-gram overlap.
For WebShop and ALFWorld, we found that the contamination may be attributed to the template-based data construction process.
For instance, in WebShop, instructions consistently followed specific formats like ``I would like <product> that is <size> and is the color <color>, and price lower than <price> dollars''.
Additionally, we observed that MBPP suffers from data contamination issues across all three training sets.
After manual inspection, we determined that most of the overlap occurs in importing Python packages and commonly used code snippets, such as loops.

In summary, it can be concluded that the data contamination has a minimal impact on the experimental results.
While some overlap exists between the held-in tasks and the training set, this is primarily a result of their data construction process.
Moreover, by adhering to the original train-test split of the datasets, the extent of performance overestimation is reduced.
Most importantly, the held-out tasks, which are used to assess the agents' generalized capabilities, do not suffer from the issue of data contamination. 
This ensures the trustworthiness and robustness of our evaluation.

\section{Complete Experimental Results}
\label{app:main}

Table \ref{tab:heldin-results} shows the complete results on held-in tasks.

\begin{table*}[t]
\centering
\resizebox{\textwidth}{!}{
\begin{tabular}{l|ccccccccc|c}
\toprule

\multirow{2}{*}{\textbf{Model}} & \multicolumn{10}{c}{\textbf{Held-in Tasks}} \\
\cmidrule(l){2-11}
& HotpotQA & StrategyQA & GSM8K & MATH & IC-SQL & MBPP & Mind2Web & WebShop & ALFWorld & Avg. \\
\midrule

\multicolumn{11}{c}{\textit{Closed-Source Model}} \\
\midrule
GPT-4   & 52.1 & 71.0 & 87.0 & 59.0 & 37.8 & 72.0 & 22.6 & 58.6 & 77.8 & 59.8 \\ 
GPT-3.5-Turbo  & 24.0 & 58.0 & 65.0 & 18.0 & 38.5 & 64.0 & 21.7 & 62.4 & 10.5 & 40.2 \\ 
\midrule

\multicolumn{11}{c}{\textit{7B Open-Source Model}} \\
\midrule
Llama-2-7B-Chat  & 3.0 & 5.0 & 15.0 & 0.0 & 4.0 & 1.0 & 11.9 & 15.8 & 0.0 & 6.2 \\ 
Vicuna-7B  & 11.0 & 47.0 & 1.0 & 3.0 & 17.3 & 21.0 & 14.8 & 33.5 & 6.0 & 17.2 \\ 
CodeLlama-7B  & 2.0 & 5.0 & 7.0 & 0.0 & 3.0 & 0.0 & 17.0 & 32.5 & 0.0 & 7.4 \\ 
AgentLM-7B  & 10.0 & 49.0 & 14.0 & 6.0 & 13.9 & 10.0 & 10.6 & 63.7 & 63.4 & 26.7 \\ 
\midrule
\model{}-7B  & 30.0 & 66.0 & 43.0 & 18.0 & 59.2 & 24.0 & 12.2 & 60.5 & 61.2 & 41.6 \\ 
\midrule

\multicolumn{11}{c}{\textit{13B Open-Source Model}} \\
\midrule
Llama-2-13B-Chat  & 6.0 & 19.0 & 18.0 & 3.0 & 3.0 & 13.4 & 17.2 & 5.3 & 0.0 & 9.4 \\   
Vicuna-13B  & 15.0 & 36.0 & 9.0 & 4.0 & 37.0 & 23.7 & 15.2 & 53.3 & 2.2 & 21.7 \\ 
CodeLlama-13B  & 7.0 & 20.0 & 29.0 & 8.1 & 3.0 & 7.2 & 7.6 & 23.0 & 0.0 & 11.7 \\ 
AgentLM-13B  & 24.0 & 52.0 & 21.0 & 6.1 & 25.7 & 20.0 & 11.1 & 65.0 & 52.2 & 30.8 \\ 
\midrule
\model{}-13B & 41.0 & 68.0 & 53.0 & 24.0 & 67.7 & 43.0 & 18.6 & 63.1 & 72.4 & 50.1 \\ 

\bottomrule
\end{tabular}
}
\caption{Performance of \model{} and baseline LLMs on held-in tasks.}
\label{tab:heldin-results}
\end{table*}

\section{Evaluation on AgentBench}
\label{app:agentbench}

AgentBench~\citep{liu2023agentbench} is another evaluation benchmark for LLM agents, encompassing 8 agent tasks.
However, it is worth noting that some tasks in AgentBench are already covered by \dataset{}, and some tasks may pose a risk of data contamination with our dataset.
Nevertheless, to provide a comprehensive perspective, we have included the results of \model{} on AgentBench as a point of reference in Table \ref{tab:agentbench}.

\begin{table*}[t]
\centering
\tabcolsep=13pt
\resizebox{\linewidth}{!}{
\begin{tabular}{lccccccccc}
\toprule

\multirow{2}{*}{\textbf{Model}} & \multicolumn{3}{c}{\textbf{Code-grounded}} & \multicolumn{3}{c}{\textbf{Game-grounded}} & \multicolumn{2}{c}{\textbf{Web-grounded}} & \multirow{2}{*}{\textbf{Overall}} \\
\cmidrule(l){2-4} \cmidrule(l){5-7} \cmidrule(l){8-9} 
& OS$^\dag$ & DB$^\dag$ & KG$^\dag$ & DCG & LTP & HH$^\ddag$ & WS$^\ddag$ & WB$^\ddag$ & \\
\midrule
GPT-4   & 42.4 & 32.0 & 58.8 & 74.5 & 16.6 & 78.0 & 61.1 & 29.0 & 4.01 \\ 
GPT-3.5-Turbo  & 32.6 & 36.7 & 25.9 & 33.7 & 10.5 & 16.0 & 64.1 & 20.0 & 2.32 \\ 
\midrule
Llama-2-7B-Chat  & 4.2 & 8.0 & 2.1 & 6.9 & 0.0 & 0.0 & 11.6 & 7.0 & 0.34 \\ 
Vicuna-7B  & 9.7 & 8.7 & 2.5 & 0.3 & 6.4 & 0.0 & 2.2 & 9.0 & 0.56 \\ 
CodeLlama-7B  & 4.9 & 12.7 & 8.2 & 0.0 & 0.0 & 2.0 & 25.2 & 12.0 & 0.50 \\ 
\midrule
\model{}-7B  & 11.8 & 9.7 & 2.7 & 1.9 & 8.2 & 68.0 & 60.5 & 12.2 & 1.60 \\ 
\bottomrule
\end{tabular}
}
\caption{Performance of \model{} and baseline LLMs on AgentBench~\citep{liu2023agentbench}. $\dag$ means the test set may suffer data contamination with \dataset{}. $\ddag$ means the task is already covered by \dataset{}.}
\label{tab:agentbench}
\end{table*}

\onecolumn

\section{Prompts for Trajectory Annotation}
\label{app:prompt_ann}

We provide the prompts for \dataset{} annotation, including answer forcing, trajectory reformat, and rationale generation.

\begin{tcolorbox}[breakable, title=Prompt for answer forcing, enhanced jigsaw]
\columnseprule=0.5pt
You are a helpful assistant. You should interact with the environment step-by-step and solve the task. I will give you some useful information to help you solve the task: a failed trajectory, and the gold answer of the task. Please solve the task again and avoid to make the same error.\\
\\
Task description:\\
\{task\_desc\}\\
\\
Failed Trajectory:\\
\{orig\_traj\}\\
\\
The correct answer of the task:\\
\{gold\_ans\}\\
\\
You have to think and solve the problem step-by-step with interleaving Thought, Action, Observation steps. At each turn, you should first provide your step-by-step thinking for solving the task. Then give your action for current step.
When you think the problem has been solved, you should give the final answer, like "Thought: your thought. Final Answer: the final answer"
\end{tcolorbox}

\begin{tcolorbox}[breakable, title=Prompt for trajectory reformat, enhanced jigsaw]
\columnseprule=0.5pt
Please help me do some reformatting work. I will give you the math question and the answer with the thinking process. Please reformat it to "Think, Act, Observation" style.\\
\\
Here is an example:\\
\\
Question: Janet's ducks lay 16 eggs per day. She eats three for breakfast every morning and bakes muffins for her friends every day with four. She sells the remainder at the farmers' market daily for \$2 per fresh duck egg. How much in dollars does she make every day at the farmers' market?\\
\\
Original format:\\
Thought process: Janet sells 16 - 3 - 4 = 9 duck eggs a day. She makes 9 * 2 = 18 every day at the farmer's market.\\
Answer: 18\\
\\
After reformatting:\\
Thought: First, I should calculate the number of duck eggs Janet sells a day\\
Action: 16 - 3 - 4\\
Observation: 9\\
Thought: Now, I should calculate the amount of money Janet makes every day at the farmer\u2019s market\\
Action: 9 * 2\\
Observation: 18\\
Final Answer: 18\\
\\
Now it's your turn. Please reformat the following question and answer.\\
\\
Question: \{question\}\\
\\
Original format:\\
Thought process: \{thought\}\\
Answer: \{answer\}\\
\\
After reformatting:\\
\end{tcolorbox}

\begin{tcolorbox}[breakable, title=Prompt for rationale generation, enhanced jigsaw]
\columnseprule=0.5pt
You are a helpful assistant. Please help me add thought process to the given trajectory.\\
\\
Task description:\\
\{task\_desc\}\\
\\
Original Trajectory:\\
\{orig\_traj\}\\
\\
You should generate the corresponding thought process which following the format:\\
Thought 1: xxx\\
Action 1: xxx\\
\\
Thought 2: xxx\\
Action 2: xxx\\
\\
Don't be lazy! DO NOT skip any actions, even if the action is repeated! Generate the corresponding thought process for each action.\\
\\
The corresponding thought process:\\
\end{tcolorbox}

\twocolumn

\end{document}